\documentclass{article}

\usepackage{arxiv}

\usepackage[utf8]{inputenc} 
\usepackage[T1]{fontenc}    
\usepackage{hyperref}       
\usepackage{url}            
\usepackage{booktabs}       
\usepackage{amsfonts}       
\usepackage{nicefrac}       
\usepackage{microtype}      
\usepackage{lipsum}		
\usepackage{graphicx}
\usepackage{natbib}
\usepackage{doi}

\usepackage{outlines}

\usepackage{multicol}
\usepackage{booktabs}
\usepackage{tabularx}

\usepackage{threeparttable}
\usepackage{multirow}
\usepackage{makecell}

\usepackage{color}

\usepackage{array}
\newcommand{\PreserveBackslash}[1]{\let\temp=\\#1\let\\=\temp}

\newcolumntype{C}[1]{>{\PreserveBackslash\centering}p{#1}}
\newcolumntype{R}[1]{>{\PreserveBackslash\raggedleft}p{#1}}
\newcolumntype{L}[1]{>{\PreserveBackslash\raggedright}p{#1}}

\usepackage{amsmath,amssymb,amsfonts}
\usepackage{graphicx}
\usepackage{textcomp}
\usepackage{xcolor}

\usepackage{siunitx}
\usepackage{bm}

\usepackage{float}
\usepackage{caption}
\usepackage{subcaption}

\newcommand{\bftab}{\fontseries{b}\selectfont}

\newcommand{\RNum}[1]{\uppercase\expandafter{\romannumeral #1\relax}}

\usepackage{algorithm}%
\usepackage{algorithmicx}%
\usepackage{algpseudocode}%

\title{Explaining Neural Networks without Access to Training Data}

\date{} 					

\author{ \href{https://orcid.org/0000-0001-8151-9223}{\includegraphics[scale=0.06]{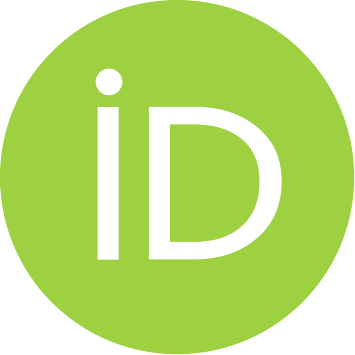}\hspace{1mm}Sascha Marton}\\
	Institute for Enterprise Systems\\
	University of Mannheim\\
	Mannheim, 68161 \\
	\texttt{marton@es.uni-mannheim.de} \\
	\And
	\href{https://orcid.org/0000-0002-1488-4236}{\includegraphics[scale=0.06]{orcid.pdf}\hspace{1mm}Stefan Lüdtke} \\
	Institute for Enterprise Systems\\
	University of Mannheim\\
	Mannheim, 68161 \\
	\texttt{luedtke@es.uni-mannheim.de} \\
	\And
	\href{https://orcid.org/0000-0003-0426-6714}{\includegraphics[scale=0.06]{orcid.pdf}\hspace{1mm}Christian Bartelt} \\
	Institute for Enterprise Systems\\
	University of Mannheim\\
	Mannheim, 68161 \\
	\texttt{bartelt@es.uni-mannheim.de} \\	
	\And
	\href{https://orcid.org/0000-0002-0638-5744}{\includegraphics[scale=0.06]{orcid.pdf}\hspace{1mm}Andrej Tschalzev} \\
	Institute for Enterprise Systems\\
	University of Mannheim\\
	Mannheim, 68161 \\
	\texttt{tschalzev@es.uni-mannheim.de} \\	
	\And
	\href{https://orcid.org/0000-0002-0209-3859}{\includegraphics[scale=0.06]{orcid.pdf}\hspace{1mm}Heiner Stuckenschmidt} \\
	Data and Web Science Group\\
	University of Mannheim\\
	Mannheim, 68159 \\
	\texttt{heiner@informatik.uni-mannheim.de} \\
}

\renewcommand{\shorttitle}{Explaining Neural Networks without Access to Training Data}

\hypersetup{
pdftitle={Explaining Neural Networks without Access to Training Data},
pdfsubject={cs.AI},
pdfauthor={Sascha Marton, Stefan Lüdtke, Christian Bartelt, Andrej Tschalzev, Heiner Stuckenschmidt},
pdfkeywords={Explainable AI (XAI), Interpretability, Neural Networks, Decision Trees},
}

\begin{document}
\maketitle

\begin{abstract}
We consider generating explanations for neural networks in cases where the network's training data is not accessible, for instance due to privacy or safety issues. 
Recently, \mbox{$\mathcal{I}$-Nets} have been proposed as a sample-free approach to post-hoc, global model interpretability that does not require access to training data.
They formulate interpretation as a machine learning task that maps network representations (parameters) to a representation of an interpretable function.
In this paper, we extend the \mbox{$\mathcal{I}$-Net} framework to the cases of standard and soft decision trees as surrogate models. We propose a suitable decision tree representation and design of the corresponding \mbox{$\mathcal{I}$-Net} output layers. Furthermore, we make \mbox{$\mathcal{I}$-Nets} applicable to real-world tasks by considering more realistic distributions when generating the \mbox{$\mathcal{I}$-Net's} training data.
We empirically evaluate our approach against traditional global, post-hoc interpretability approaches and show that it achieves superior results when the training data is not accessible.

\end{abstract}


\section{Introduction}\label{sec:introduction}





Artificial neural networks achieve impressive results for various modeling tasks~\citep{lecun2015deep, wang2020recent}.
However, a downside of their superior performance and sophisticated structure is the comprehensibility of the learned models. 
In many domains, it is crucial to understand the function learned by a neural network, especially when it comes to decisions that affect people~\citep{samek2019, molnar2020interpretable}. 


A common approach to tackle the problem of  interpretability without sacrificing the superior performance is using a surrogate model as gateway to interpretability \citep{molnar2020interpretable}. 
Most existing global surrogate approaches use a distillation procedure to learn the surrogate model based on the predictions of the neural network~\citep{molnar2020interpretable, frosst2017distilling}. Therefore, they query the neural network based on a representative set of samples and the resulting input-output pairs are then used to train the surrogate model. This representative sample usually comprises the training data of the original model, or at least follows its distribution~\citep{molnar2020interpretable, lopes2017data}. However, there are many cases where the training data cannot easily be exposed due to privacy or safety concerns~\citep{lopes2017data,bhardwaj2019dream,nayak2019zero}.
Without having access to the training data, traditional approaches can fail to provide meaningful explanations since the querying strategy can easily miss dense regions of the training data such that the resulting samples are a poor approximation of the true function, as we will show in the following example. \\

\emph{Example 1.} 
The Credit Card Default dataset~\citep{dataset_credit_card} comprises personal, confidential data which usually cannot be exposed to external authorities. 
The task is to predict whether a client will default the payment in the current month, which can be solved efficiently using neural networks. 
To gain insight into the decision-making process of the neural network, we can learn a global surrogate model.
Unfortunately, if the training data is not accessible, we can't ensure that the neural network is properly queried, and the explanation contains the relevant information when using a traditional, sample-based distillation.

\begin{figure}[htb]
\centering
\begin{subfigure}{.5\columnwidth}
   \centering
  \includegraphics[width=\columnwidth]{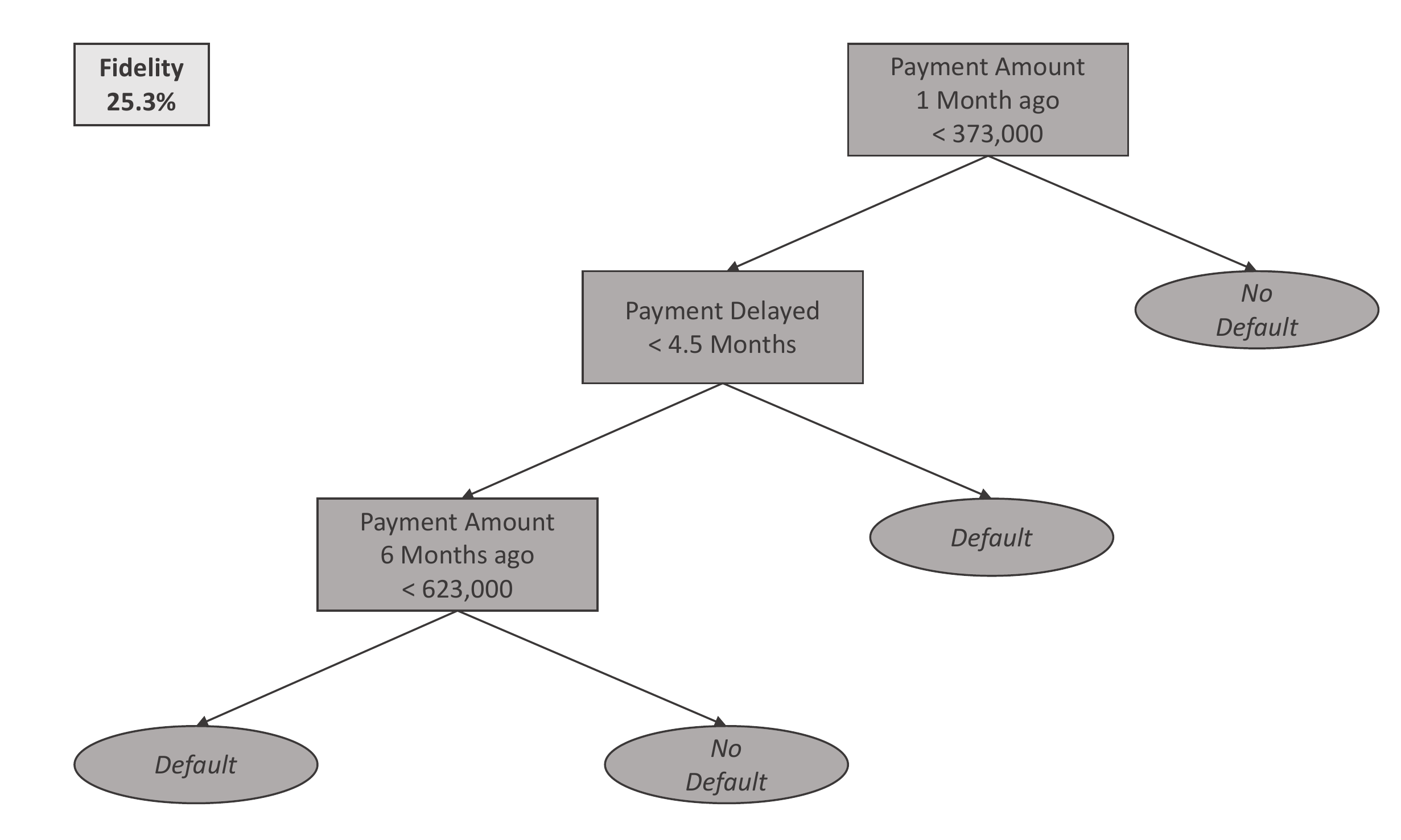}
  \caption{Sample-Based Decision Tree}
  \label{fig:vanilla_credit_card_reduced_sample}
  
\end{subfigure}%
\begin{subfigure}{.5\columnwidth}
  \centering
  \includegraphics[width=\columnwidth]{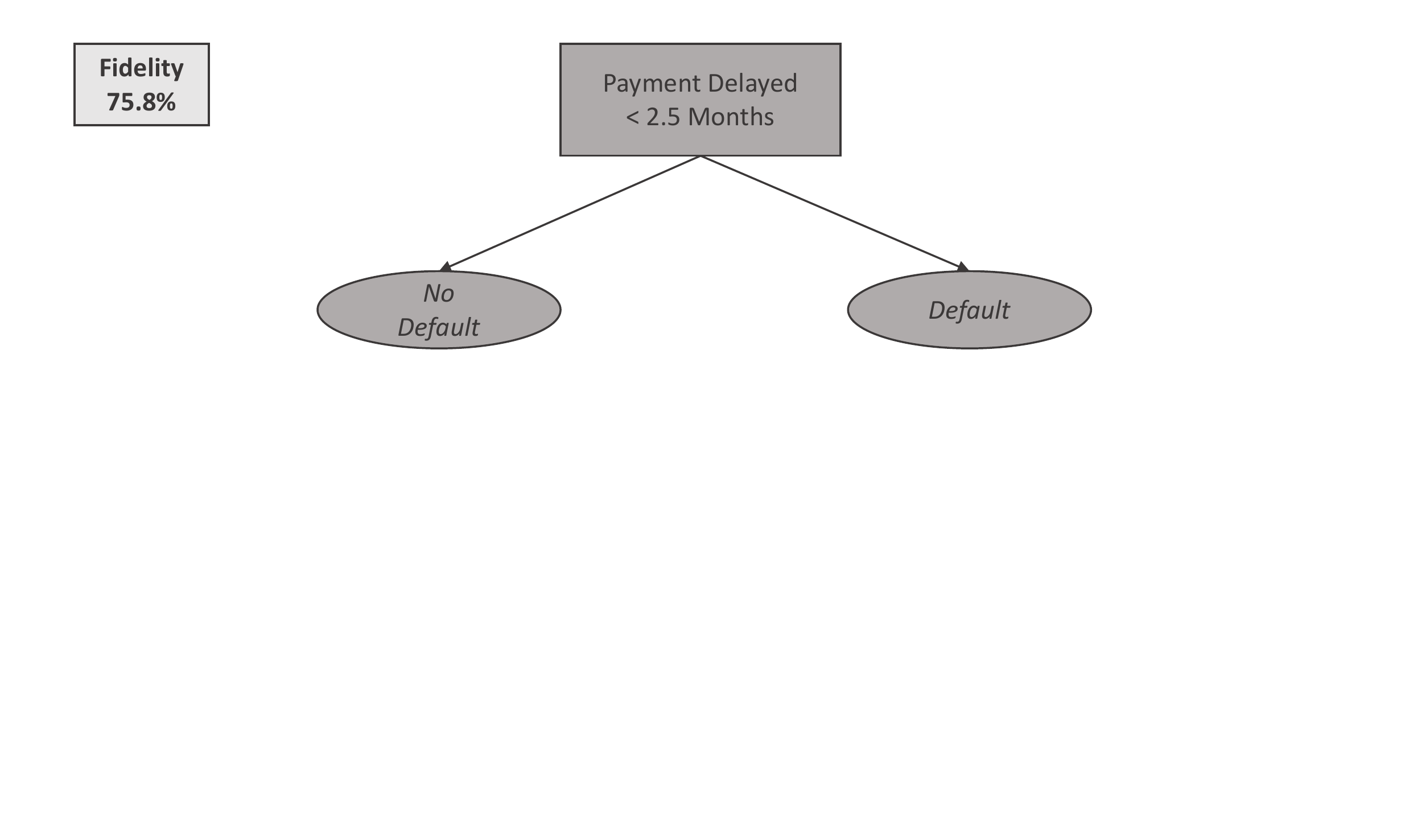}
  \caption{\mbox{$\mathcal{I}$-Net} Decision Tree}
  \label{fig:vanilla_credit_card_reduced_inet}
\end{subfigure}
\caption{\textbf{Explaining Neural Networks for Credit Card Default Prediction.} The DT on the left is learned by a sample-based distillation without access to training data, and the DT on the right is predicted by the \mbox{$\mathcal{I}$-Net}. The \mbox{$\mathcal{I}$-Net} makes reasonable splits and achieves a significantly higher fidelity on the real data.}
\label{fig:vanilla_credit_card_reduced}
\end{figure}

Figure~\ref{fig:vanilla_credit_card_reduced_sample} shows such a scenario, where the explanation generated by a sample-based distillation without training data contains a misconception: It encodes the rule that we should always predict \emph{No Default} if the payment amount of the last month is larger than $373,000$ without taking the payment history of the client into account. 
This mismatch between network and surrogate model is also reflected in the low fidelity between the network and surrogate model on the training data. However, in reality, this fidelity cannot be computed when the training data is not available, such that these misconceptions might go unnoticed. This can lead to wrong assumptions about what the network actually learned. \\

As shown in \emph{Example 1}, knowing the training data is crucial for sample-based methods and without them, it is often not possible to generate reasonable explanations.
Recent approaches tackle this issue by using only a subset of the training data and/or layer activations to generate a representative set of samples~\citep{lopes2017data,bhardwaj2019dream,nayak2019zero}. However, they still rely on a proper querying of the model and use a sample-based distillation.

In contrast, the \mbox{$\mathcal{I}$-Net} approach introduced by \citet{marton2022explanations} is a sample-free approach that only accesses the network parameters and therefore does not rely on a proper querying. 
This is achieved by using a second neural network (the so-called \mbox{$\mathcal{I}$-Net}) which learns a mapping from the network parameters to a human-understandable representation of the network function. 
Following this approach, we can generate reasonable explanations, even when the training data is not accessible, as shown in Figure~\ref{fig:vanilla_credit_card_reduced_inet}: The \mbox{$\mathcal{I}$-Net} achieves a high fidelity and the resulting surrogate model encodes the rule that \emph{Default} is predicted if the payment for the last month was delayed for more than two months, which is a reasonable explanation for this scenario.


The \mbox{$\mathcal{I}$-Net} was originally devised for lower-order polynomials as a surrogate model. While polynomials can be reasonable explanations for regression tasks, they are not well-suited for representing decision boundaries of a classification task, which is the focus of our paper. In contrast, decision trees (DTs) are frequently used as an explainable model for classification tasks since they make hierarchical decisions and therefore are easy to comprehend for humans~\citep{frosst2017distilling}. 
In the recent literature, soft DTs (SDTs) are successfully used as interpretable surrogate models~\citep{frosst2017distilling}. While SDTs make multivariate splits, they usually achieve a higher fidelity than standard DTs, but also have a higher level of complexity. 

In this paper, we make the following contributions:
\begin{itemize}
    \item We propose an improved data generation method for the training of an \mbox{$\mathcal{I}$-Net} (Section~\ref{sssec:data_generation}) which allows more robust and distribution-independent explanations on real-world datasets.
    \item We present an \mbox{$\mathcal{I}$-Net} design that is able to represent standard DTs and SDTs as surrogate model (Section~\ref{ssec:inets_dts}) with a high fidelity.
    \item We introduce univariate SDTs as complexity-fidelity trade-off (Section~\ref{sssec:inets_dts_soft}) and show that they can be implemented efficiently using the \mbox{$\mathcal{I}$-Net}.
    \item We empirically evaluate our approach against sample-based approaches for learning standard DTs and SDTs and show that it achieves superior results when training data is not accessible (Section~\ref{ssec:exp_results}).
\end{itemize}


\section{\mbox{$\mathcal{I}$-Nets} as a Sample-Free Approach to Global Model Interpretability}\label{sec:inets}
In this section, we summarize the task of explaining neural networks, focusing on the case where the networks' training data is not available, followed by a brief introduction to the \mbox{$\mathcal{I}$-Net} approach. For a more in-depth explanation of the \mbox{$\mathcal{I}$-Net} approach, we refer to \citet{marton2022explanations}.

\subsection{Global Explanations for Neural Networks}\label{ssec:explaining_nn}

The general task of globally explaining neural networks can be formalized as finding a function $g: X \rightarrow P(Y \lvert X)$ (i.e., a surrogate model) that approximates the decision function of a neural network  $\lambda: X \rightarrow P(Y \lvert X)$, such that $\forall \mathbf{x} \in X: \lambda(\mathbf{x}) \approx g(\mathbf{x})$, where $X$ is a set of feature vectors and $Y$ is a set of classes.

Since the \mbox{$\mathcal{I}$-Net} approach implements a learning task, it is convenient to distinguish between the functions $\lambda$ and $g$ and their representations $\theta_\lambda \in \Theta_\lambda$ and $\theta_g \in \Theta_g$~\citep{marton2022explanations}. The representation $\theta_\lambda$ consists of the network parameters, i.e., the weights and biases of the neural network. Similarly, $\theta_g$ is the parameter vector of the surrogate model and depends on the selected function family. 

\begin{figure}[tb]
    \includegraphics[width=0.99\columnwidth]{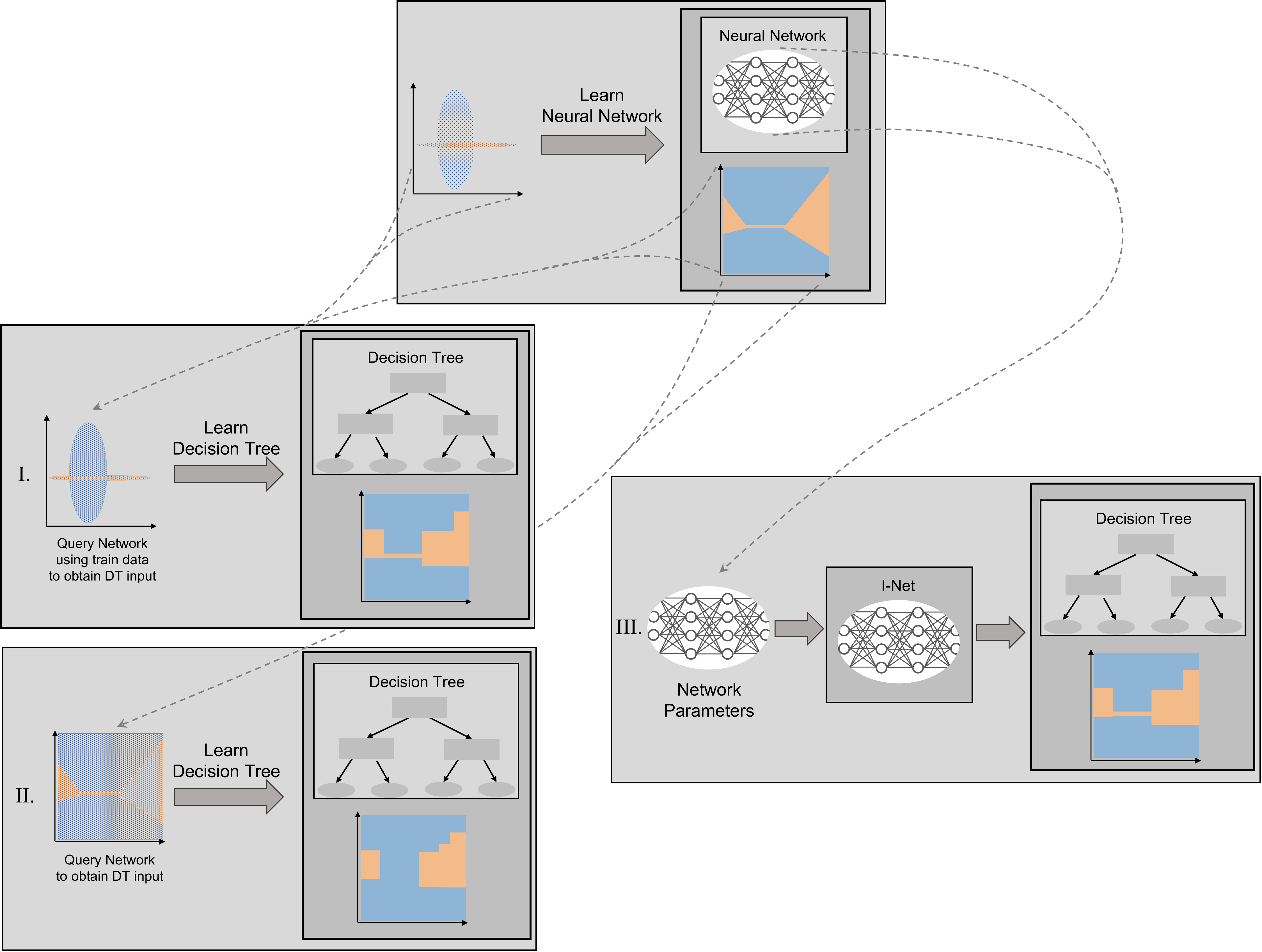}
    \caption{\textbf{Sample-Based and $\boldsymbol{\mathcal{I}}$-Net Approach.} Sample-based approaches query the target network based on a set of data points. Using the train data to query the network (\RNum{1}) usually generates a meaningful explanation. If the training data is not available, we have to query the network based on randomly sampled data, e.g., from a uniform distribution (\RNum{2}), which often cannot generate a meaningful explanation since relevant parts are not queried properly. The \mbox{$\mathcal{I}$-Net} uses the network parameters as an input to generate a reasonable explanation (\RNum{3}) and does not rely on querying the neural network.}

    \label{fig:decision-boundary-explanation}
\end{figure}

The process of generating explanations can be formalized as a function $\mathcal{I}: \Theta_\lambda \rightarrow \Theta_g$ that maps representations of $\lambda$ to representations of $g$~\citep{marton2022explanations}. 
Traditional approaches for generating global surrogate models post-hoc implement $\mathcal{I}$ via a sample-based procedure, as shown in Figure~\ref{fig:decision-boundary-explanation} (\RNum{1}-\RNum{2}). 
They generate a new dataset, where the labels are obtained by querying $\lambda$ based on a set of data points. In the next step, a surrogate model is trained using the generated dataset, maximizing the fidelity between $\lambda$ and $g$.
As shown by \citet{marton2022explanations}, this process is time-consuming, which can be a huge drawback if timely explanations are required, as for instance in an online learning scenario.
Additionally, it also strongly depends on the data used for querying the model.
Information that is not properly queried cannot be contained in the explanation, as already shown in \emph{Example 1}. Therefore, in the literature it is suggested to use the original train data or data from the same distribution for querying the model~\citep{molnar2020interpretable, lopes2017data}, which usually yields to meaningful explanations as depicted in \RNum{1} in Figure~\ref{fig:decision-boundary-explanation}. However, if the training data is not accessible or not existing anymore, the model has to be queried based on some sampled data (\RNum{2} in Figure~\ref{fig:decision-boundary-explanation}). 
In this case, it is often not possible to generate meaningful explanations with sample-based approaches, since we cannot ensure a proper querying and therefore the explanation does not necessarily focus on the relevant aspects.

\subsection{Reasonable Explanations}\label{ssec:reasonable_explanations}
\begin{figure}[tb]
    \centering
    \includegraphics[width=0.80\columnwidth]{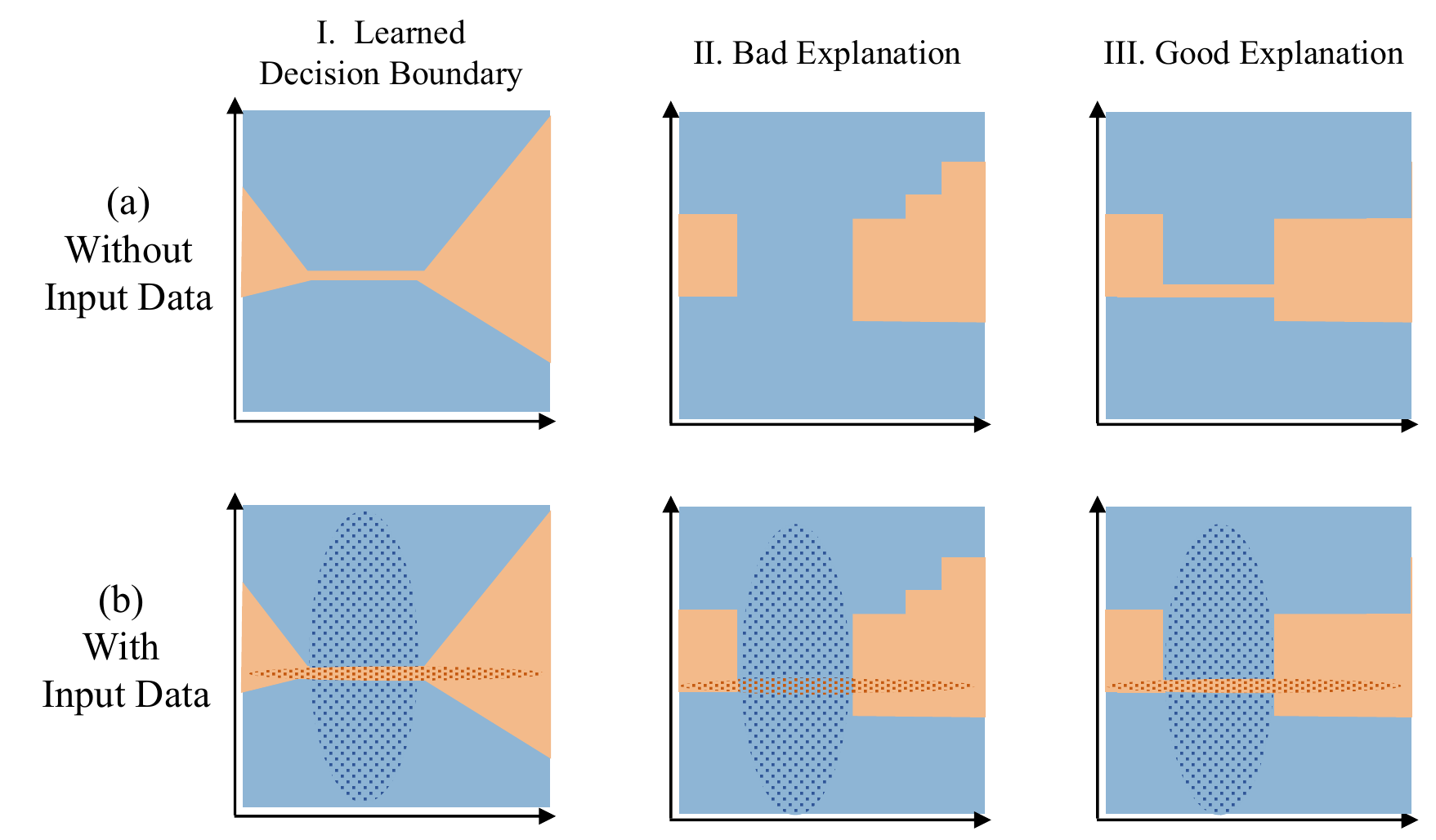}
    \caption{\textbf{Good and Bad Explanations.} This figure shows an exemplary decision boundary of a bad (\RNum{2}) and a good (\RNum{3}) explanation for the model we want to interpret (\RNum{1}). Without considering the data (a), the explanation shown in \RNum{2} appears very reasonable, since the areas created by the decision boundary cover most of the decision boundary of the original model. However, when taking the data into account (b), we can see that the small area in the center of the picture is very important, since there are many samples located. This is neglected by the explanation shown in \RNum{2} and only considered by the explanation shown in \RNum{3}.}

    \label{fig:decision-boundary-explanation_simple}
\end{figure}

In the following, we discuss what constitutes a meaningful explanation for a neural network. 
In general, the decision boundary of the surrogate model should closely match the decision boundary of the network we want to interpret to achieve a high fidelity. However, we argue that it is necessary to also take the data distribution into account: A decision boundary should assign as many samples as possible to the correct class. Therefore, it is crucial that the decision boundary is composed correctly in the areas where many samples are located.
Accordingly, for a reasonable explanation, the decision boundary should match the model we want to interpret especially in regions where many samples are located, while it is less important that the decision boundaries match in regions with low data density.
In other words, we are less interested in an explanation that shows us how the model behaves when making predictions on data points that do not occur in reality.
This concept is visualized in Figure~\ref{fig:decision-boundary-explanation_simple}.
In Section~\ref{sssec:visual_eval}, we show that traditional, sample-based approaches cannot generate such reasonable explanations when the training data is not available.



\subsection{Explanations for Neural Networks by Neural Networks}
\begin{figure}[tb]
    \includegraphics[width=0.99\columnwidth]{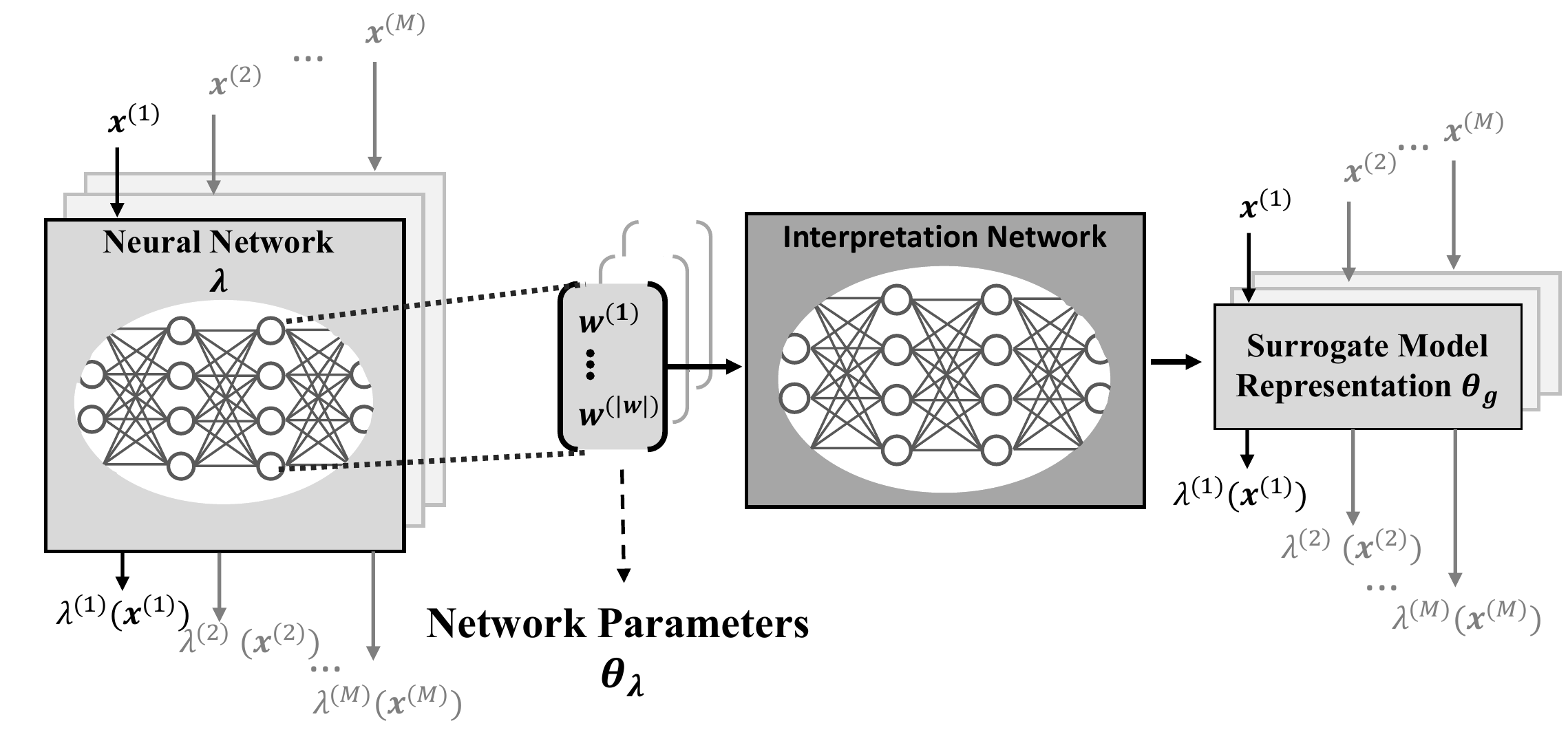}
    \caption{\textbf{Overview of the \mbox{$\boldsymbol{\mathcal{I}}$-Net}  Approach.} The neural network parameters as input are translated into a surrogate model~\cite{marton2022explanations}.}

    \label{fig:overview}
\end{figure}

To renounce the dependency on a proper querying of the model, we can implement $\mathcal{I}$ as a neural network as proposed by \citet{marton2022explanations}. Therefore, we transform the task of explaining neural networks into a machine learning task, as shown in Figure~\ref{fig:decision-boundary-explanation}~\RNum{3}.
The concept of \mbox{$\mathcal{I}$-Nets} is depicted more detailed in Figure~\ref{fig:overview} and involves two major steps: 
\begin{enumerate}
    \item We train a set of neural networks on synthetic data and extract their learned parameters.
    \item We train a second neural network, the \mbox{$\mathcal{I}$-Net}, using the parameters extracted in the first step as input data.
\end{enumerate}

Thereby, no supervision in terms of actual labels is required during the training. Instead, the fidelity between $\lambda$ and $g$ is computed using a distance measure over a set of data points in the loss function. Since the loss is only computed during the training, no data except the network parameters is required when applying the \mbox{$\mathcal{I}$-Net}. This is a major advantage to sample-based approaches, where the training data is required for each network we want to interpret.
Accordingly, to generate an explanation, \mbox{$\mathcal{I}$-Nets} only need access to the network parameters and therefore, the approach can be applied in scenarios where the training data is not accessible without suffering a performance deficit. 

The most crucial part of the \mbox{$\mathcal{I}$-Net} approach is an efficient training procedure. Thereby, as for most machine learning tasks, good training data (in our case, a set of network parameters $\Theta_\lambda$) is important. 
Therefore, we present an improved data generation method making \mbox{$\mathcal{I}$-Nets} applicable for real world scenarios in Section~\ref{sssec:data_generation}.

\section{Robust \mbox{$\mathcal{I}$-Nets} for Decision Trees}\label{ssec:inets_improvements}

In this section, we present the main contributions of this paper. 
\citet{marton2022explanations} argue that \mbox{$\mathcal{I}$-Nets} can be trained solely based on synthetic data. However, it is crucial that this synthetic data comprises reasonable learning problems to assure that an application of the \mbox{$\mathcal{I}$-Net} is possible in a real-world setting. Therefore, we will introduce an improved data generation method that considers multiple data distributions and creates reasonable learning tasks (Section~\ref{sssec:data_generation}). 
Furthermore, \citet{marton2022explanations} focus on regression tasks. In contrast, we focus on classification tasks and therefore present an adjusted loss function for the \mbox{$\mathcal{I}$-Net} (Section~\ref{sssec:loss}).

In general, the \mbox{$\mathcal{I}$-Net} framework can be applied to arbitrary function families for $g$, as long as a suitable representation $\theta_g$ is available. 
In Section~\ref{ssec:inets_dts}, we introduce different DT variants and propose corresponding representations $\theta_g$ that allow an efficient training.


\subsection{Improved Data Generation and Training Procedure}\label{sssec:improvements}

\subsubsection{Data Generation Method}\label{sssec:data_generation}
The data generation method proposed by \citet{marton2022explanations} focused on maximizing the performance of the \mbox{$\mathcal{I}$-Net} during training by learning functions $\lambda$ that are similar to the function family of $g$. This is achieved by randomly sampling a set of functions from the family of $g$. These functions are queried to generate labels for a uniformly sampled dataset, which is used to learn $\lambda$.
This procedure ensures that the functions $\lambda$ are representative of $g$, which enables efficient training.
However, a high training performance does not necessarily mean that the model generalizes well to unseen data, i.e., neural networks trained on real-world datasets. 

Additionally, \citet{marton2022explanations} use a uniform data distribution to query $\lambda$ for the fidelity calculation in the \mbox{$\mathcal{I}$-Net} loss. 
However, if we only consider a uniform distribution during the training of the \mbox{$\mathcal{I}$-Net}, we might not be able to make reasonable predictions if the network we want to interpret was trained using data from a substantially different distribution, as we will show in Section~\ref{sssec:improvements_eval}.
This problem is related to the general problem that occurs for a machine learning task, if the data we are actually interested in (i.e., the test data) is from a different distribution than the data used for training the model.

To tackle this issue, we propose using multiple, different distributions during the training of the \mbox{$\mathcal{I}$-Net} to make it more robust and therefore applicable on real-world datasets. In this process, we can also utilize the fact that an \mbox{$\mathcal{I}$-Net} can be trained in a controlled, synthetic environment~\citep{marton2022explanations}: For each $\theta_\lambda \in \Theta_\lambda$, we know the data that was used for learning $\lambda$. Therefore, we can use these data points to compute the \mbox{$\mathcal{I}$-Net} loss on a meaningful set of samples during the training. The \mbox{$\mathcal{I}$-Net} utilizes this additional knowledge to generalize. Since the loss is only calculated during training, it can generate meaningful explanations solely based on the network parameters $\theta_\lambda$ at test time.


In general, generating the training data for the \mbox{$\mathcal{I}$-Net} $\Theta_\lambda$ involves three major steps:
\begin{enumerate}
    \item Generate $N$ datasets $\mathcal{D}_\lambda = \{(\mathbf{x}^{(j)},y^{(j)})\}_{j=1}^M$, each comprising $M$ samples.
    \item For each dataset $\mathcal{D}_\lambda$, train a network $\lambda$, extract the network parameters $\theta_\lambda$ and add them to the training set $\Theta_\lambda$.
    \item Use $\Theta_\lambda$ to train an \mbox{$\mathcal{I}$-Net} for the respective function family.
\end{enumerate}

\begin{table}[tb]
\centering
\begin{tabular}{cccc}
\toprule
\textbf{Distribution} & \textbf{Parameter} $\boldsymbol{p_1}$ & \textbf{Parameter} $\boldsymbol{p_2}$ & \textbf{Symbol} \\ \midrule
Uniform & minimum & maximum & $\mathcal{U}(p_1, p_2)$ \\
Normal & location (mean) & scale (standard deviation) & $\mathcal{N}(p_1, p_2)$ \\
Gamma & shape ($k$) & scale ($\theta$) & $\Gamma(p_1, p_2)$  \\
Beta & $\alpha$ & $\beta$ & $\text{B}(p_1, p_2)$  \\ 
Poisson & lambda & - & $\text{Poi}(p_1)$ \\ \bottomrule
\end{tabular}
\caption{\textbf{Data Generation Distributions.} This table summarizes the different distributions used for the data generation of the \mbox{$\mathcal{I}$-Net} with their parameters and used symbol.}
\label{tab:distrib}
\end{table}

\begin{figure}[tb]
    \includegraphics[width=0.99\columnwidth]{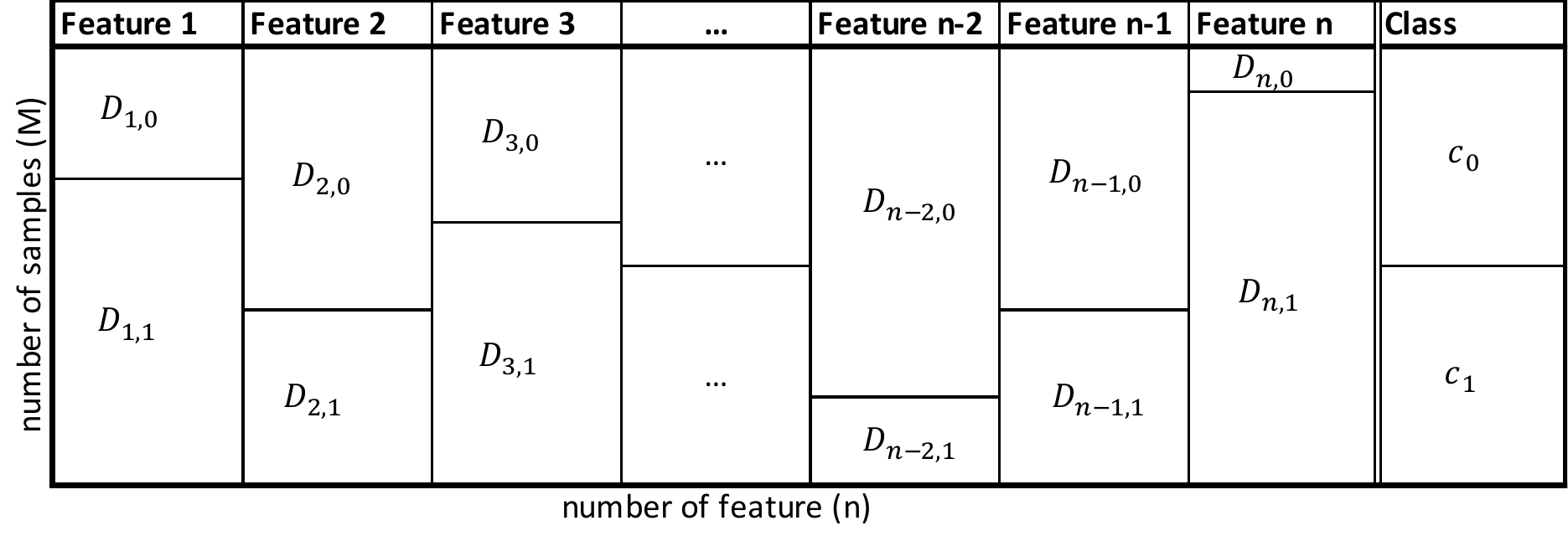}
    \caption{\textbf{Data Generation Visualization.} This graphic visualizes the generation of a balanced, random dataset used for training a network $\lambda$ where $D \in \{\mathcal{U}, \mathcal{N}, \Gamma, \text{B}, \text{Poi}\}$. For each feature, a random distribution with two random parametrizations is chosen and a random number of data points is sampled from each distribution.}
    \label{fig:distrib}
\end{figure}

The data generation is visualized in Figure~\ref{fig:distrib}:
For each feature $i$, we sample data points from one distribution with $k$ different parametrizations, where $k$ is the number of classes. For this paper, we focus on binary classification tasks and therefore set $k=2$. The distribution $D_{i,k}$ is sampled uniformly from $\{\mathcal{U}, \mathcal{N}, \Gamma, \text{B}, \text{Poi}\}$ for each feature. The distributions considered within this paper are summarized in Table~\ref{tab:distrib}. The distributions were selected to cover a wide range of diverse distributions that are reasonable for many different real-world phenomena~\citep{distribution_relation, mun2015understanding}.
The parametrization for the distributions $D_{i,0}$ and $D_{i,1}$ are again randomly drawn from $\mathcal{U}(0,p)$, where $p$ is a hyperparameter for the data generation procedure.
The number of samples is selected randomly, where $M_0 =  \lceil \mathcal{U}(1,M-1) \rceil$ data points are sampled from $D_{i,0}$ and  $M_1 = M - M_0$ data points are sampled from $D_{i,1}$. The generated datasets are balanced and for each feature and the first $\frac{M}{2}$ data points are associated with \emph{Class 0} and the subsequent $\frac{M}{2}$ data points are associated with \emph{Class 1}. This procedure is formalized in Algorithm~\ref{alg:generate-data}. 
We can see the proposed data generation method as a generalization of common, synthetic machine learning problems (as for instance make\_blobs\footnote{Accessible using sklearn under \url{https://scikit-learn.org/0.15/modules/generated/sklearn.datasets.make_blobs.html} (Accessed 15.05.2022)}), that is able to generate more realistic tasks.

\begin{algorithm}[tb]
\caption{Generate Multi-Distribution Data}   \label{alg:generate-data} 

\begin{algorithmic}[1] 
    \Function{generate}{$n,D,M$}
            \For{$i = 1,\dots,n$}
                \State{$D_i \sim \mathcal{U}\{\mathcal{U}, \mathcal{N}, \Gamma, \text{B}, \text{Poi}\}$} \Comment{Randomly chose distribution}
                \State{$M_0 \sim \lceil \mathcal{U}(1,M-1) \rceil$}
                \State{$\mathbf{p_0} \sim \mathcal{U}(0,p)$} \Comment{Randomly sample parameters for $D_i,0$}
                \State{$\mathbf{p_1} \sim \mathcal{U}(0,p)$} \Comment{Randomly sample parameters for $D_i,1$}
                \For{$j = 1,\dots,M_0$}
                    \State{$x_i^{(k)} \sim D_i(\mathbf{p_0})$}
                \EndFor
                \For{$j = 1,\dots,M-M_0$}
                    \State{$x_{i}^{(M_0+j)} \sim D_i(\mathbf{p_1})$}         
                \EndFor
                
                \State{$\mathbf{x}_i \leftarrow \frac{\mathbf{x}_i - min(\mathbf{x}_i)}{max(\mathbf{x}_i) - min(\mathbf{x}_i})$}
                \Comment{Scale feature to $[0,1]$} 
   	        \EndFor

            \For{$j = 1,\dots, \lceil \frac{M}{2} \rceil$}
   	            \State{$y^{(j)} \leftarrow 0$}
            \EndFor
            \For{$j = 1,\dots, \lfloor \frac{M}{2} \rfloor$}
   	            \State{$y^{(j)} \leftarrow 1$}
            \EndFor   
            \State $\mathcal{D} \leftarrow \{\mathbf{x}^{(j)},y^{(j)}\}_{j=1}^{M}$
            \State \Return{$\mathcal{D}$}
    \EndFunction
\end{algorithmic}

\end{algorithm}

\subsubsection{Adjusted Loss Function}\label{sssec:loss}
While \citet{marton2022explanations} focused on regression tasks, we focus on binary classification tasks within this paper.
Therefore, the loss function has to be adjusted by using binary cross-entropy as distance measure to quantify the fidelity between $\lambda$ and $g$:

\begin{equation}\label{eq:bc_inet}
\begin{split}
    \text{BC}(\theta_\lambda,\theta_g) = \frac{1}{M} \sum_{j=1}^M & \lfloor \lambda(\mathbf{x}^{(j)}) \rceil \times  log(g(\mathbf{x}^{(j)}) ) \\
     + & (1 - \lfloor \lambda(\mathbf{x}^{(j)}) \rceil) \times log(1 - g(\mathbf{x}^{(j)}) )    
\end{split}
\end{equation}

Here, $\lfloor \cdot \rceil$ denotes rounding to the closest integer, which is required to calculate the binary cross-entropy correctly.
The \mbox{$\mathcal{I}$-Net} loss for a set of network parameters $\Theta_\lambda = \{\theta_\lambda^{(i)}\}_{i=1}^N$ is then computed as

\begin{equation}\label{eq:loss_inet}
    \mathcal{L}_{\mathcal{I}} = \frac{1}{\lvert \Theta_\lambda \rvert} \sum_{\theta_\lambda \in \Theta_\lambda} \text{BC}(\theta_\lambda,\mathcal{I}(\theta_\lambda)).
\end{equation}

\subsection{Function Families and \mbox{$\mathcal{I}$-Net} Output Representation}\label{ssec:inets_dts}

\subsubsection{\mbox{$\mathcal{I}$-Nets} for Standard Decision Trees}\label{sssec:inets_dts_standard}
The first function family we will consider as surrogate models are standard DTs. DTs and decision rules are frequently used as explanations, since they are comparatively easy to understand for most humans~\citep{molnar2020interpretable}. 

\begin{figure}[tb]
    \centering
    \includegraphics[width=0.8\columnwidth]{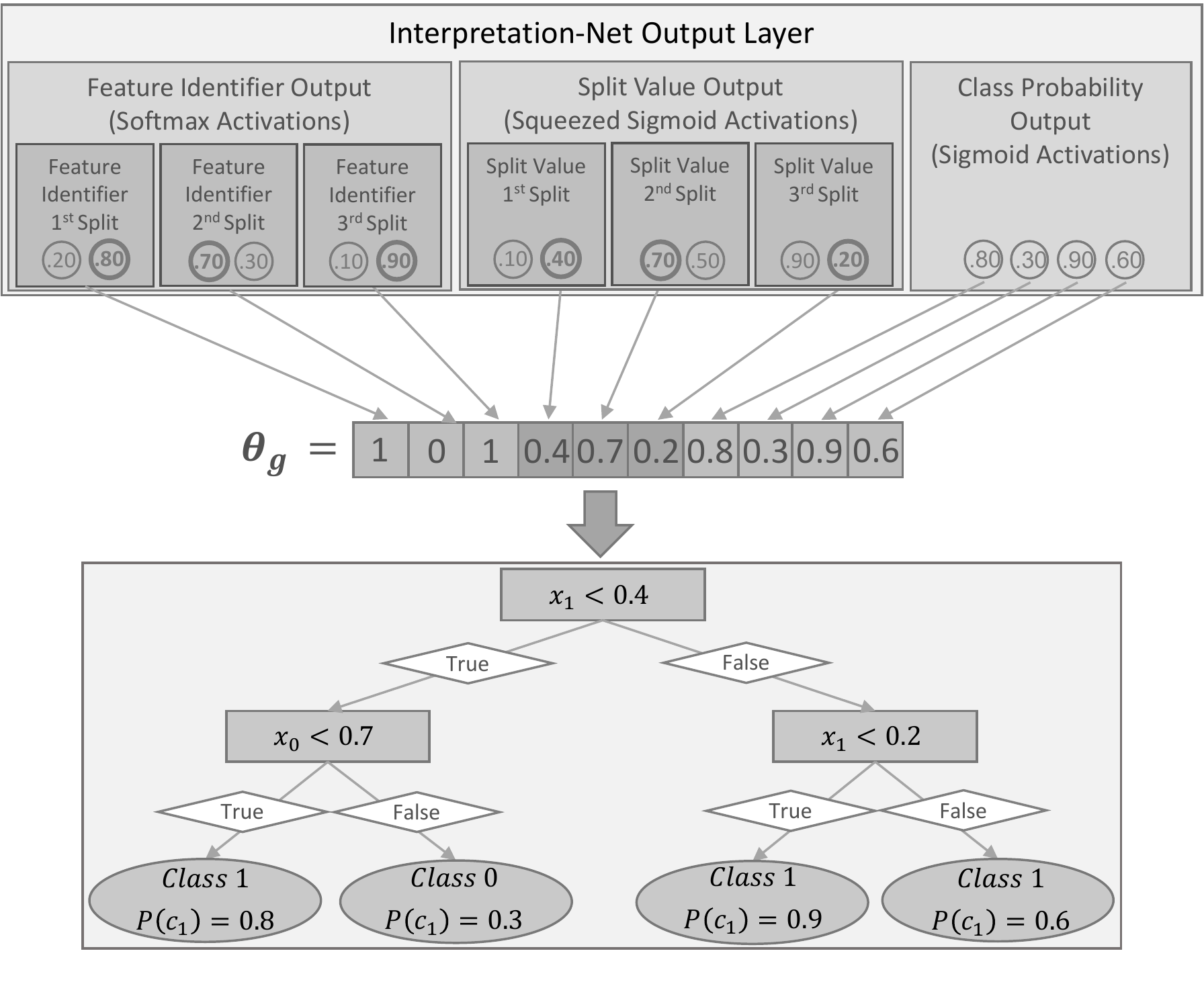}
    \caption{\textbf{Exemplary $\boldsymbol{\mathcal{I}}$-Net Output for DTs.} The DT representation is predicted by the \mbox{$\mathcal{I}$-Net} using three separate output layers with different activation functions. The output shows an exemplary DT of depth two for a binary classification task on a dataset with two features.}
    \label{fig:inet_dt}
\end{figure}

\paragraph{Standard Decision Tree Representation in the \mbox{$\mathcal{I}$-Net} Framework}
The \mbox{$\mathcal{I}$-Net} approach requires a suitable representation $\theta_g$ for standard DTs to enable efficient learning. Specifically, we need a one-dimensional encoding of internal and leaf nodes, as shown in Figure~\ref{fig:inet_dt}.

The inner node of a DT comprises two major parts: The first part is the feature that is considered within the split, and the second part is the split value. The operator is fixed to less ($<$) as it is common practice for representing DTs. Furthermore, we can fix the left path to be the true path and the right path as the false path. 

The feature $x_i$ considered in the split can be defined by enumerating the features, where $i \in \{0,1, \dots ,n\}$. We can represent this using $n$ neurons and a softmax activation for each inner node (i.e., we can see it as a classification task for which feature to consider at a certain split).

For the split value, we can assume that all features are scaled to be within $[0, 1]$, as it is common practice. To represent this in the \mbox{$\mathcal{I}$-Net} output, we can use sigmoid activations, to constraint the output interval. However, due to the functional form of the sigmoid activation, the \mbox{$\mathcal{I}$-Net} prefers split values close to $0.5$. To counteract this tendency, we used a squeezed sigmoid activation, which we define as $\frac{1}{1+e^{-3x}}$. This supports the \mbox{$\mathcal{I}$-Net} in choosing more distinct split values. 
Furthermore, the output layer does not comprise one split value for each split, but $n$ split values for each split (one for each feature). To construct the DT, we always use the split value at the index indicated by the feature identifier. This design choice is influenced by the fact that we always need to consider the meaning of a split value in context with the corresponding feature. In other words, while the split value $0.7$ might be a reasonable threshold for the feature $x_0$, it might not be reasonable at all for the feature $x_1$. Designing the \mbox{$\mathcal{I}$-Net} output with one split value for each feature and each inner node, we can make this interaction easier to learn.

In a standard DT, the leaf nodes comprise the decision for a certain path (i.e., the class to be predicted). However, to compute the \mbox{$\mathcal{I}$-Net} loss in Equation~\ref{eq:loss_inet}, it is necessary that $g$ has probabilities as an output. Therefore, we adjust the DTs to not just have a class in the leaf node, but a probability. This is similar to the purity in the leaf node of a DT, which is also often used as a gateway to predicting probabilities using a standard DT. In a binary classification case, we can use a single value to represent the probabilities of predicting \emph{Class 1} and thereby, the probability of \emph{Class 0} is the complementary probability. In the output layer of the \mbox{$\mathcal{I}$-Net}, we can represent this using a total of $2^d$ neurons with sigmoid activations (one neuron for each leaf node).
This can easily be extended for a multi-class classification problem with $k$ classes by using $k \times 2^d$ neurons and one softmax activation over $k$ neurons (one softmax for each leaf node).

\subsubsection{\mbox{$\mathcal{I}$-Nets} for Soft Decision Trees}\label{sssec:inets_dts_soft}

SDTs were proposed by \citet{frosst2017distilling} to overcome the interpretability problem that arises from distributed hierarchical representations when using neural networks by expressing the knowledge using hierarchical decisions of a tree-based structure.
Unlike standard DTs, SDTs do not make hard true/false splits at each internal node, but use soft decisions associated with probabilities for each path. In the following, we will shortly introduce the functioning of SDTs. For a more in-depth description, especially concerning the learning algorithm, we refer to \citet{frosst2017distilling}.

\begin{figure}[tb]
    \centering
    \includegraphics[width=0.70\columnwidth]{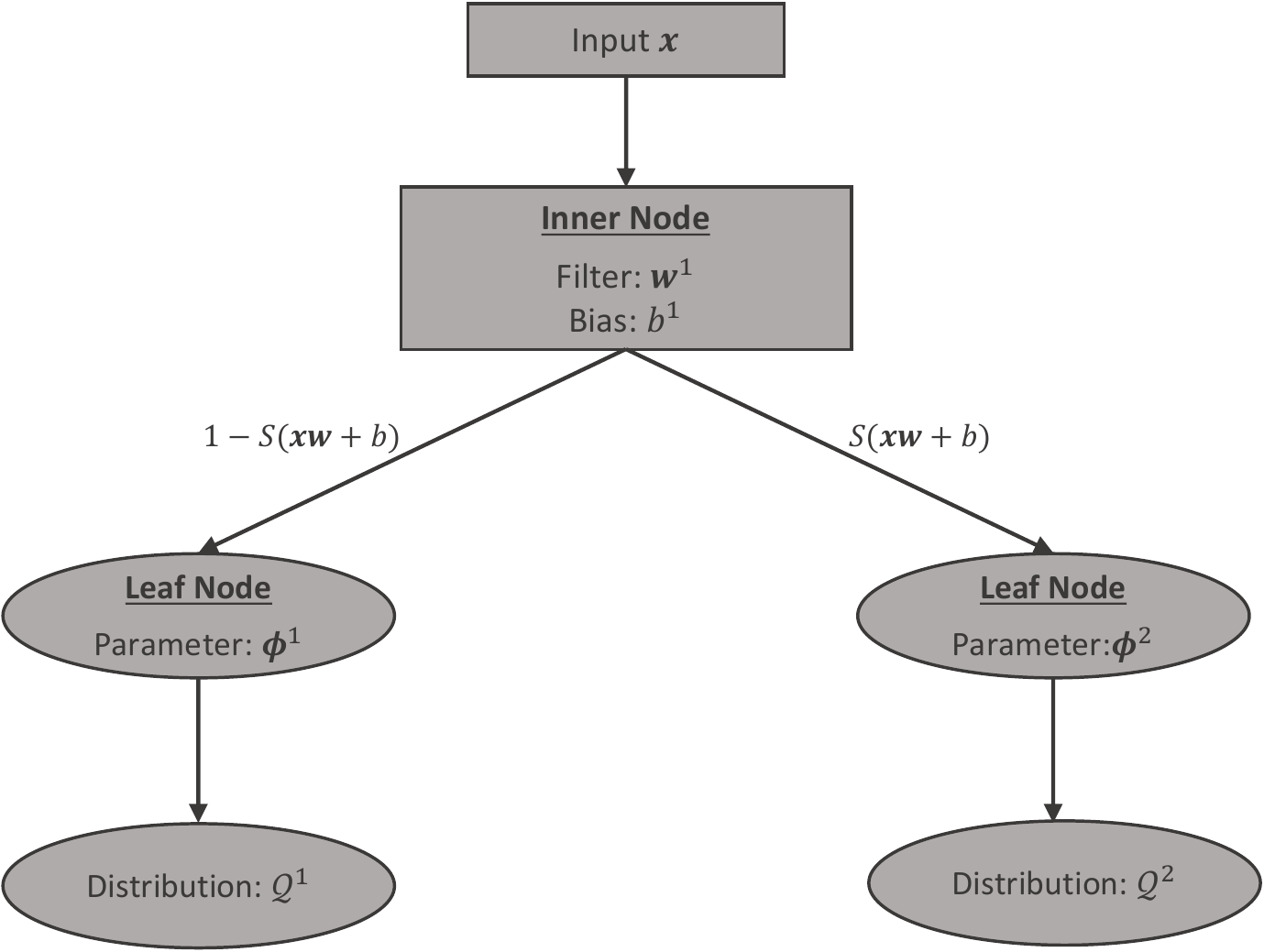}
    \caption{\textbf{Soft Decision Tree} This figure shows a minimal SDT with a single inner node and two leaf nodes.}
    \label{fig:sdt}
\end{figure}

Figure~\ref{fig:sdt} shows a SDT with a single internal node following the design of \citet{frosst2017distilling}. Each internal node $j$ comprises a filter $\mathbf{w}^j$ and a bias $b^j$. While the bias is a single, learned value, the filter consists of one value for each feature. Accordingly, in contrast to a standard DT with univariate decisions, a SDT has a multivariate decision at each internal node. This comes with a significantly higher model complexity, especially with an increasing number of features.

At each internal node, the probability of taking the right branch is calculated by 
\begin{equation}
    P^j(\mathbf{x}) = S(\mathbf{x}w^j + b^j)
\end{equation}
where $x$ is an input sample and $S$ is a sigmoid function defined as $S(x) = \frac{1}{1+e^{-x}}$.
Each leaf nodes $l$ comprise a probability distribution $Q^l_.$, which for the binary case is defined as
\begin{equation}
    Q^l_k = \frac{e^{\phi^l_k}}{e^{\phi^l_0} + e^{\phi^l_1}}
\end{equation}
where $k \in \{0,1\}$ is the output class and $\phi^l_.$ is a learned parameter for each leaf $l$.

Usually when using SDTs, there is not only a single leaf node considered when making a prediction, but all leaf nodes are multiplied with their path probabilities to calculate the final probability distribution. However, \citet{frosst2017distilling} suggest increasing the interpretability of SDTs by just considering the path with the maximum path probability when calculating the final probability distribution. Since this does not significantly affect the performance, we will only consider SDTs using the maximum path probability in this paper.

While \citet{frosst2017distilling} focused on image datasets for a better visualization of the explanation, they also reported that their algorithm can efficiently be applied to tabular data. However, for tabular data, especially when the feature space is high-dimensional, using SDTs as a surrogate model can make the explanation hard to understand for humans due to their high complexity. To counteract the strong correlation between the model complexity and the number of features, we propose univariate SDTs. Univariate SDTs are similar to standard SDTs, but the filter at each internal node comprises only one value $\neq 0$.
While the complexity of univariate SDTs is similar to standard DTs, they can represent decision boundaries that are not parallel to the feature axis. Simultaneously, they maintain the hierarchical structure, making them easier to comprehend for humans.

Unfortunately, univariate SDTs cannot be trained using the algorithm proposed by \citet{frosst2017distilling} without adjustments and there is to the best of our knowledge no existing learning algorithm for univariate SDTs. 
However, we can constrain the learning algorithm to just consider the feature with the highest absolute value (which is also the value that according to the design has the highest influence on the decision) when calculating the path probabilities. We can implement this by multiplying the filter with a binary mask generated by an argmax or exaggerated softmax of the absolute filter values. Accordingly, during the forward pass, the filter values are calculated as:
\begin{equation}
    \mathbf{w}_i = \mathbf{w}_i \times \sigma(\beta_2 \times \lvert \mathbf{w}_i \rvert)
\end{equation}
By increasing the temperature $\beta_2$, we increase the extent to which the filter with the highest absolute value influences the calculation of the path probability. When $\beta_2$ is sufficiently high, we approximate an argmax and the path probability is calculated only based on the filter with the highest absolute value.

\paragraph{Soft Decision Tree Representations in the \mbox{$\mathcal{I}$-Net} Framework}

\begin{figure}[tb]
    \includegraphics[width=0.95\columnwidth]{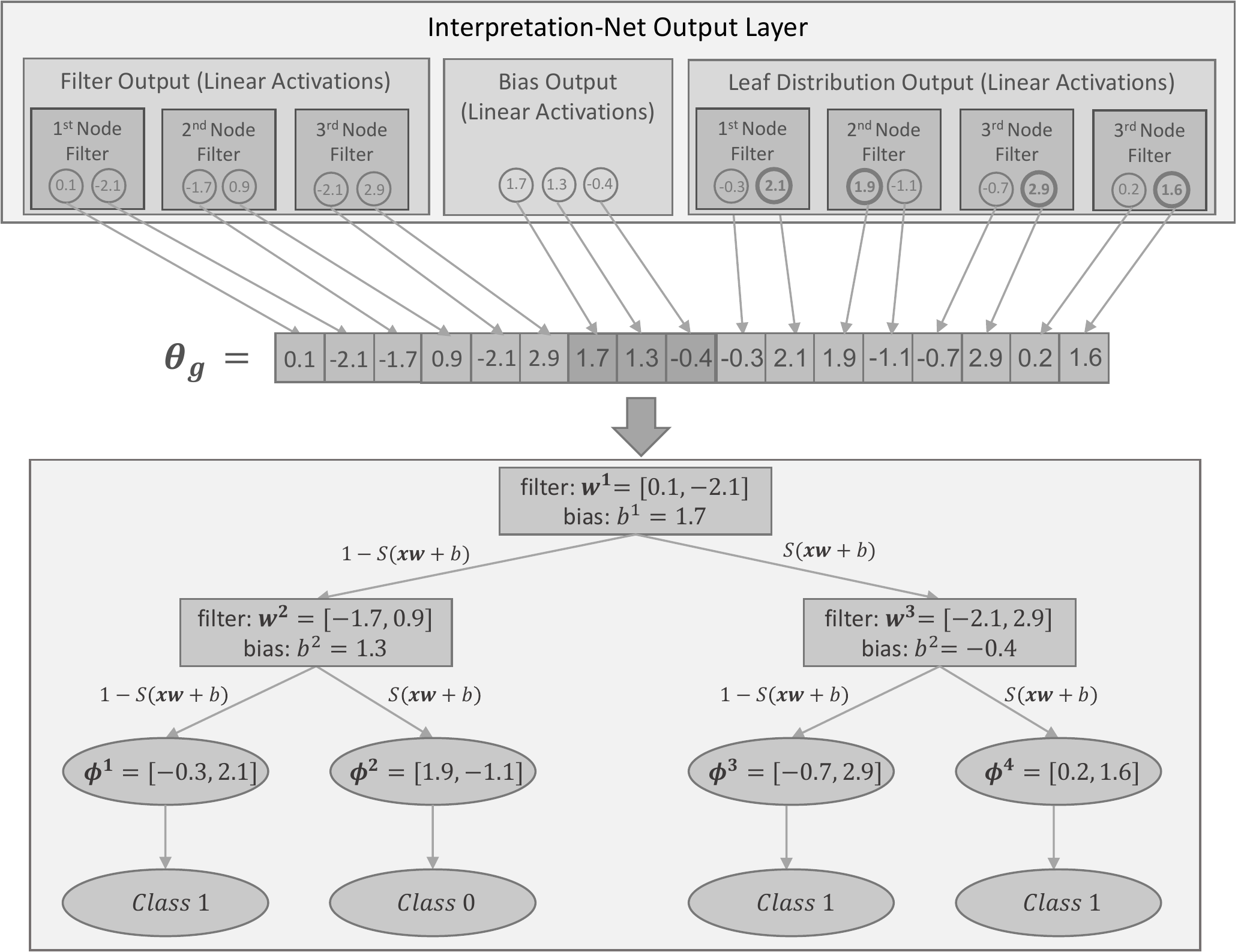}
    \caption{\textbf{Exemplary $\boldsymbol{\mathcal{I}}$-Net Output for SDTs.} The SDT representation is predicted by the \mbox{$\mathcal{I}$-Net} using three separate output layers with linear activation functions. The output shows an exemplary SDT of depth two for a binary classification task on a dataset with two features.}    
    \label{fig:inet_sdt}
\end{figure}

To use SDTs as surrogate models within the \mbox{$\mathcal{I}$-Net} framework, we again need a suitable representation $\theta_g$. Fortunately, the encoding for standard SDT shown in Figure~\ref{fig:inet_sdt} is straightforward: We can represent the internal nodes with $n$ output neurons for the filter (one for each feature) and one output neuron for the bias. Since there are no specific ranges for the filter and bias value in the SDT, we use linear activations. The same accounts for $\phi^l_.$, where we need $k$ output neurons for each leaf node. Again, we can use linear activations here, since the final probabilities are calculated by $Q^l_k$ and no specific range for $\phi^l_.$ is required. 

\begin{figure}[tb]
    \includegraphics[width=0.95\columnwidth]{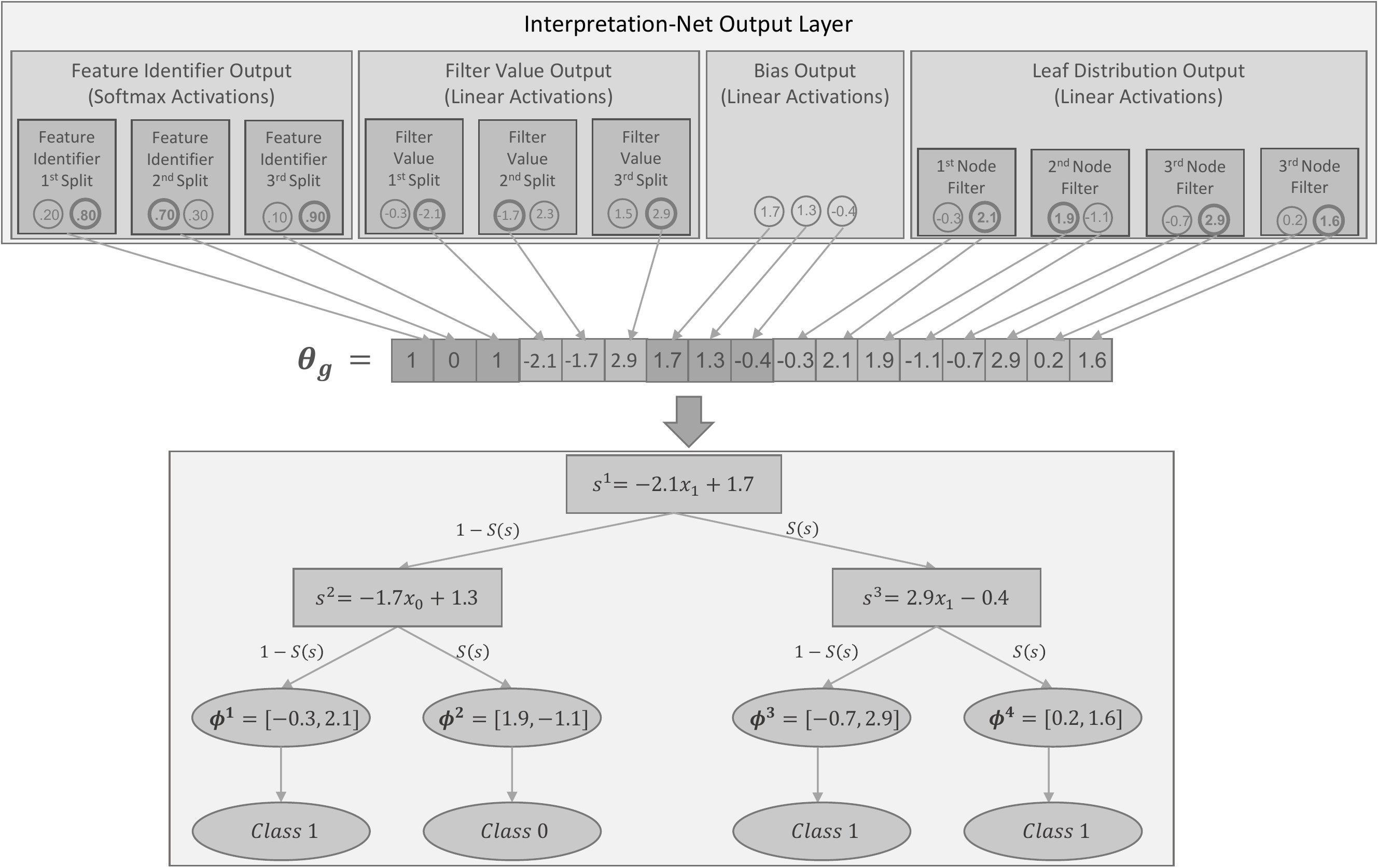}
    \caption{\textbf{Exemplary $\boldsymbol{\mathcal{I}}$-Net Output for Univariate SDTs.} The univariate SDT representation is predicted by the \mbox{$\mathcal{I}$-Net} using four separate output layers with different activation functions. The output shows an exemplary univariate SDT of depth two for a binary classification task on a dataset with two features.}    
    \label{fig:inet_sdt1}
\end{figure}

Unfortunately, the representation of univariate SDTs as \mbox{$\mathcal{I}$-Net} output is not as straightforward. This is mainly caused by the fact that neural networks hardly predict a value of $0$, which would be necessary when using the same representation we proposed for standard SDTs. This would result in many small filter values and therefore still multivariate trees. 
One solution is using a feature identifier and a separate filter value output, similar to the representation of the internal nodes for standard DTs (Figure~\ref{fig:inet_dt}).
Accordingly, we can use $n$ output neurons and a softmax activation for each internal node to indicate at which position to set the filter value (i.e., we can see this as a classification task for which filter value is $\neq 0$). We represent the filter value again using $n$ neurons with a linear activation for each internal node.
The bias and leaf parameters are adopted from the standard SDT. An exemplary representation for univariate SDTs is shown in Figure~\ref{fig:inet_sdt1}.



\section{Evaluation} \label{sec:eval}
The goal of our evaluation is to show that \mbox{$\mathcal{I}$-Nets} are able to interpret neural networks trained on real-world datasets without requiring access to the training data, and achieve a higher fidelity than sample-based approaches in most of the cases. Therefore, we will address the following in our evaluation:
\begin{itemize}
    \item We illustrate which effects occur once the training data is not accessible for querying the model and thereby show that it is crucial for sample-based approaches to access the training data (Section~\ref{sssec:visual_eval}).
    \item We empirically investigate the fidelity of the \mbox{$\mathcal{I}$-Net} in comparison to sample-based approaches on real-world datasets if the training data is not accessible (Section~\ref{sssec:real_world_eval}).
    \item We perform an ablation study showing the effect of the improved data generation method introduced in Section~\ref{sssec:data_generation} on the \mbox{$\mathcal{I}$-Net} performance for real-world datasets (Section~\ref{sssec:improvements_eval}).    
    \item We present a case study of credit card default prediction, comparing the explanations for a neural network generated by the \mbox{$\mathcal{I}$-Net} and sample-based approaches without access to the training data (Section~\ref{sssec:case_study_eval}).
\end{itemize}

\subsection{Experimental Setup}\label{ssec:experimental_setup}
Within our experiments, we compare \mbox{$\mathcal{I}$-Nets} with standard distillation approaches for a scenario where the original training data is not available.
Thereby, we used the representations $\Theta_g$ described in Section~\ref{ssec:inets_dts} for the \mbox{$\mathcal{I}$-Net}. The sample-based distillation was conducted as follows:


\begin{itemize}
    \item \textbf{Standard Decision Trees: } For standard DTs, we used the implementation from sklearn\footnote{Available under: \url{https://scikit-learn.org/stable/modules/generated/sklearn.tree.DecisionTreeClassifier.html}} (Accessed 15.05.2022) which uses the CART algorithm for DT induction~\citep{breiman1984classification}.
    \item \textbf{Univariate Soft Decision Trees: } For univariate SDTs, we used the learning algorithm introduced by \citet{frosst2017distilling} with our adjustments described in Section~\ref{sssec:inets_dts_soft}.
    \item \textbf{Standard Soft Decision Trees: } For SDTs, we used the algorithm proposed by \citet{frosst2017distilling}.
\end{itemize}
The hyperparameters used during our experiments are summarized in Table~\ref{tab:dt-params}-\ref{tab:sdt-params}. 

Since we assume that the training data is not available, we needed to generate data for querying the neural network to distill a surrogate model. Therefore, we selected three sampling strategies for generating the query data as benchmarks:
\begin{enumerate}
    \item \textbf{Multi-Distribution: } According to Algorithm~\ref{alg:generate-data}, i.e., considering different data distributions to allow for a fair comparison with the \mbox{$\mathcal{I}$-Net}.
    \item \textbf{Standard Uniform: }  A standard uniform distribution $U(0,1)$.
    \item  \textbf{Standard Normal: } A standard normal distribution $\mathcal{N}(0, 1)$.
\end{enumerate}
For each sampling strategy, we sampled $10000$ data points. However, this is only necessary for the sample-based approaches and the \mbox{$\mathcal{I}$-Net} as sample-free approach does not rely on querying to generate explanations.
Increasing the number of sampling points further did not enhance the fidelity of sample-based approaches (see Figure~\ref{fig:sample_size}), but only increases their runtime.
We further assumed, that the data used for training the neural network was scaled to be in $[0, 1]$ and therefore, we also scaled the sampled data to the same interval.

The network parameters $\Theta_\lambda$ for training the \mbox{$\mathcal{I}$-Net} were generated according to Algorithm~\ref{alg:generate-data} using the hyperparameters in Table~\ref{tab:lambda-net-params}. The hyperparameter $p$ which defines the maximum value for the distribution parameters was fixed to $5$ during the data generation for all experiments. Furthermore, we excluded all datasets that were linearly separable during the data generation to focus on more complex and reasonable datasets. 
The \mbox{$\mathcal{I}$-Net} hyperparameters are summarized in Table~\ref{tab:i-net-params} and were tuned using a greedy neural architecture search according to \citet{jin2019auto} followed by a manual fine-tuning of the selected values. We selected one \mbox{$\mathcal{I}$-Net} architecture for each of the three function families.

The neural networks considered during the evaluation in Section~\ref{ssec:exp_results} were trained with the hyperparameters summarized in Table~\ref{tab:lambda-net-params}. The fidelity between the surrogate model and the neural network is always calculated based on the test split of the original data.

\subsection{Experimental Results}\label{ssec:exp_results}

\subsubsection{Visual Evaluation for Different Distributions}\label{sssec:visual_eval}
\begin{figure}[tb]
    \includegraphics[width=0.99\columnwidth]{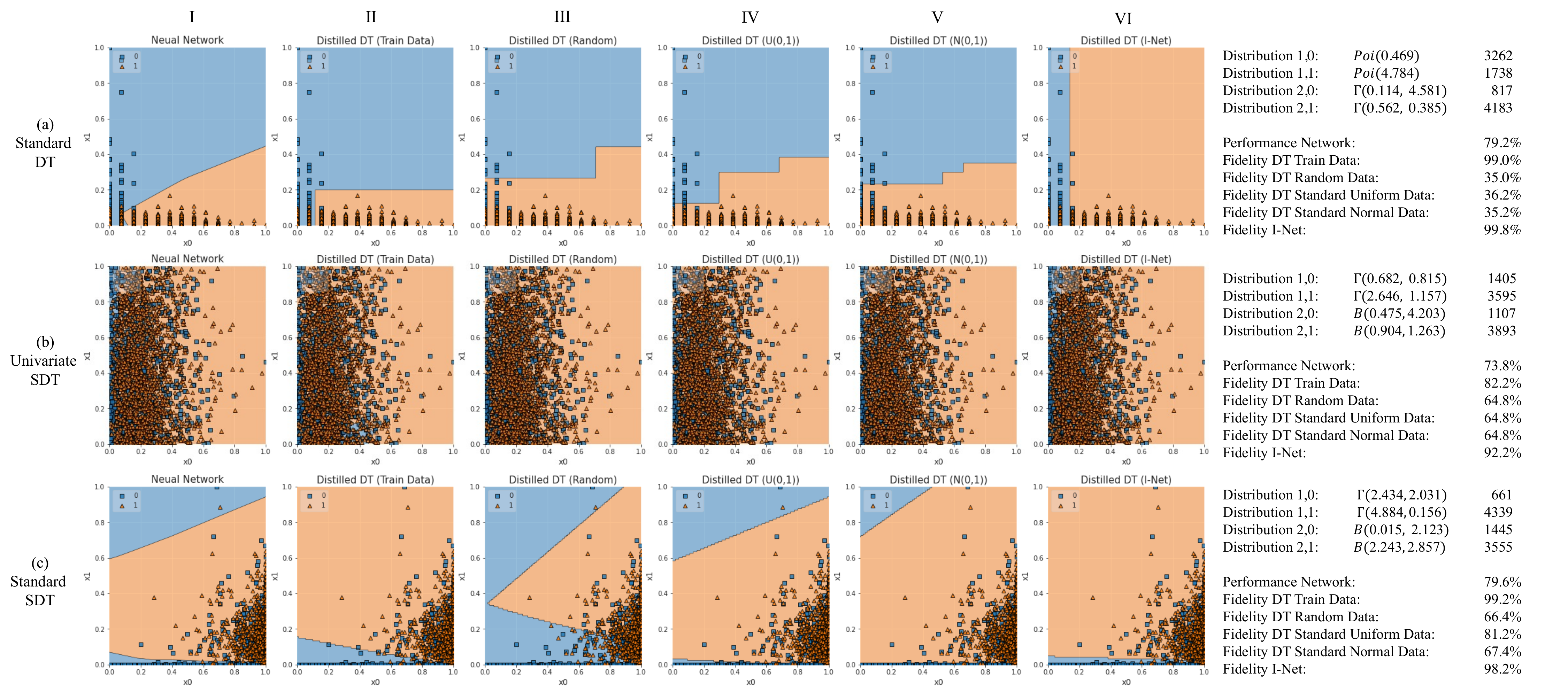}
    \caption{\textbf{Visual Decision Boundary Evaluation.} This figure shows the decision boundaries of the neural network we want to interpret (\RNum{1}), followed by the decision boundary of explanations generated by different approaches, along with their performance for three different datasets and function families. Only when the training data is accessed (\RNum{2}) and when using the \mbox{$\mathcal{I}$-Net} (\RNum{6}), the explanation comprises the relevant aspects of the model. When the training data is not accessible (\RNum{3})-(\RNum{5}), sample-based approaches are not able to generate reasonable explanations.}
    \label{fig:graph-eval-combined}
\end{figure}

In this experiment, we show the importance of knowing the distribution of the training data in a controlled, synthetic setting. We use a two-dimensional dataset which allows a visual comparison of the decision boundaries (Figure~\ref{fig:graph-eval-combined}). The data used for training the neural networks for this experiment was generated randomly according to Algorithm~\ref{alg:generate-data} but is distinct to the data used for training the \mbox{$\mathcal{I}$-Net} to ensure a fair comparison. 


Figure~\ref{fig:graph-eval-combined}(a) shows a decision boundary learned by a neural network that ranges from the bottom left corner to the middle right. Thereby, many data points that were assigned to \emph{Class 0} by the neural network are located in the bottom left corner. In contrast, the top right part contains few to no data points.
When the training data is available (\RNum{2}), the standard DT learned a decision boundary that closely matches the decision boundary of the neural network, including the area in the bottom left. However, if the training data is not available, the sample-based approach (\RNum{3}-\RNum{5}) only comprises the large area towards the top and neglects the small area at the bottom left. Considering just the shapes and size of the areas created by the decision boundary, this seems to be a reasonable explanation. However, as explained in Section~\ref{ssec:reasonable_explanations}, if we take the data into account, it becomes apparent that the neglected part of the decision boundary at the bottom left is much more important, since many data points are located in this area.
In contrast, the explanation generated by the \mbox{$\mathcal{I}$-Net} as sample-free approach (\RNum{6}) correctly separates the samples at the bottom left with its decision boundary and neglects the part at the top right, which is not relevant when taking the data into account.
We can confirm this by taking the fidelity scores into account: The \mbox{$\mathcal{I}$-Net} achieved a fidelity of $99.8\%$, while the sample-based distillation without training data only achieved a maximum fidelity of $36.2\%$.

For univariate and standard SDTs we can observe similar results: The fidelity for the sample-based approaches (\RNum{3})-(\RNum{5}) significantly decreased if the training data is not available and the explanation focused on non-relevant areas. In contrast, the \mbox{$\mathcal{I}$-Net} (\RNum{6}) was able to generate high-fidelity explanations without accessing the training data.

\subsubsection{Real World Datasets Performance Comparison}\label{sssec:real_world_eval}
In this experiment, we compare the performance of the \mbox{$\mathcal{I}$-Net} and a sample-based distillation without access to train data on real-world datasets. We selected $8$ commonly used datasets, mostly focusing on the banking and medical domain, comprising personal, confidential data where it is realistic to assume that the training data cannot be exposed.
A description of the datasets, the preprocessing that was performed, and the performance of the neural networks we want to interpret is summarized in Appendix~\ref{A:real-world-data}.

Table~\ref{tab:eval-results-vanilla}-\ref{tab:eval-results-sdt} show the results on the selected datasets for the \mbox{$\mathcal{I}$-Net} and a sample-based distillation using the data generation methods introduced in Section~\ref{ssec:experimental_setup} for the different function families. 
For the sample-based distillation, the results show the mean and standard deviation over $10$ trials. While for the standard uniform and standard normal sampling only the sampled data points differ, we sampled a new set of distributions and parameters for each trial in the multi-distribution case. 
For a better visual comparison, we highlighted a value bold if it shows a significantly higher fidelity for a certain dataset with $95\%$ confidence according to an unpaired t-test\footnote{For the t-test calculation, we only compared the \mbox{$\mathcal{I}$-Net} with the distillation approach for the sampling strategy with the highest fidelity. Therefore, a bold value only means that a sampling strategy was significantly better than the \mbox{$\mathcal{I}$-Net} and not that it was significantly better than the other sampling strategies. A bold value for the \mbox{$\mathcal{I}$-Net} means that it achieved a significantly higher accuracy than the best sampling strategy for the respective dataset.}. 

\paragraph{Standard Decision Trees}
Comparing the results for standard DTs as surrogate model in Table~\ref{tab:eval-results-vanilla}, the \mbox{$\mathcal{I}$-Net} had a significantly higher fidelity on $5/8$ datasets. A sample-based distillation achieved the highest fidelity on $2/8$ datasets. 
Even though the multi-distribution sampling strategy did not achieve the best performance on any dataset, it had the highest average performance among the sample-based distillation methods, with a mean fidelity of $73.2\%$. Nevertheless, the performance of the \mbox{$\mathcal{I}$-Net} was significantly better and achieved a mean fidelity of $86.19\%$. Especially for the \emph{Breast Cancer Wisconsin Original} and \emph{Wisconsin Diagnostic Breast Cancer}, the sample-based distillation was not able to generate accurate explanations if the training data was not accessible. For those datasets, the fidelity of sample-based distillation was often even worse than a random guess, which, as already shown in Section~\ref{sssec:visual_eval}, highlights the importance of querying the model on reasonable data points.


\begin{table}[tb]
\centering
\resizebox{\columnwidth}{!}{
\begin{tabular}{l|cccc}
\toprule
\textbf{Dataset} &                                          \mbox{\textbf{\boldsymbol{$\mathcal{I}$}-Net}} & \textbf{Multi-Distribution} &                         \textbf{Standard Uniform} &                                \textbf{Standard Normal} \\
\midrule
\textbf{Titanic (n=9)                            } &  \bftab\phantom{0}95.51 $\pm$ \phantom{0}0.00 &                      \phantom{0}71.12 $\pm$ 17.16 &        \phantom{0}86.07 $\pm$ \phantom{0}3.30 &        \phantom{0}86.29 $\pm$ \phantom{0}7.75 \\
\textbf{Medical Insurance (n=9)                  } &        \phantom{0}82.71 $\pm$ \phantom{0}0.00 &            \phantom{0}88.12 $\pm$ \phantom{0}6.71 &        \phantom{0}89.47 $\pm$ \phantom{0}4.19 &  \bftab\phantom{0}90.75 $\pm$ \phantom{0}8.83 \\
\textbf{Breast Cancer Wisconsin Original (n=9)             } &  \bftab\phantom{0}97.10 $\pm$ \phantom{0}0.00 &                      \phantom{0}83.62 $\pm$ 13.09 &                  \phantom{0}39.42 $\pm$ 13.90 &        \phantom{0}31.88 $\pm$ \phantom{0}0.00 \\
\textbf{Wisconsin Diagnostic Breast Cancer (n=10)} &  \bftab\phantom{0}80.36 $\pm$ \phantom{0}0.00 &                      \phantom{0}56.43 $\pm$ 17.65 &                  \phantom{0}37.86 $\pm$ 15.56 &        \phantom{0}33.39 $\pm$ \phantom{0}5.42 \\
\textbf{Heart Disease (n=13)                     } &        \phantom{0}73.33 $\pm$ \phantom{0}0.00 &            \phantom{0}74.67 $\pm$ \phantom{0}9.45 &  \bftab\phantom{0}85.67 $\pm$ \phantom{0}5.97 &        \phantom{0}80.33 $\pm$ \phantom{0}7.67 \\
\textbf{Cervical Cancer (n=15)                   } &  \bftab\phantom{0}84.71 $\pm$ \phantom{0}0.00 &                      \phantom{0}65.41 $\pm$ 27.77 &        \phantom{0}71.88 $\pm$ \phantom{0}9.64 &                  \phantom{0}60.82 $\pm$ 30.29 \\
\textbf{Loan House (n=16)                        } &                  100.00 $\pm$ \phantom{0}0.00 &                      \phantom{0}77.05 $\pm$ 24.41 &        \phantom{0}96.89 $\pm$ \phantom{0}7.42 &                  \phantom{0}59.84 $\pm$ 33.84 \\
\textbf{Credit Card Default (n=23)                       } &  \bftab\phantom{0}75.80 $\pm$ \phantom{0}0.00 &                      \phantom{0}69.16 $\pm$ 17.58 &        \phantom{0}74.76 $\pm$ \phantom{0}0.05 &                  \phantom{0}34.33 $\pm$ 20.31 \\
\midrule
\textbf{Mean Fidelity                               } &        \phantom{0}86.19 &            \phantom{0}73.20 &                  \phantom{0}72.75  &                  \phantom{0}59.70  \\
\bottomrule
\end{tabular}
}
\caption{\textbf{Real-World Evaluation Results for Standard Decision Trees.}}
\label{tab:eval-results-vanilla}
\end{table}

\paragraph{Soft Decision Trees}

For univariate SDTs, the performance difference between \mbox{$\mathcal{I}$-Nets} and sample-based methods is even more pronounced: \mbox{$\mathcal{I}$-Nets} achieved a significantly higher fidelity in $7/8$ cases with a mean fidelity of $82.3\%$, while the best sampling strategy only achieved a mean fidelity of $51.02\%$.

\begin{table}[tb]
\centering
\resizebox{\columnwidth}{!}{
\begin{tabular}{l|cccc}
\toprule
\textbf{Dataset} &                                          \mbox{\textbf{\boldsymbol{$\mathcal{I}$}-Net}} & \textbf{Multi-Distribution} &                         \textbf{Standard Uniform} &                                \textbf{Standard Normal} \\
\midrule
\textbf{Titanic (n=9)                            } &  \bftab\phantom{0}95.51 $\pm$ \phantom{0}0.00 &                      \phantom{0}47.75 $\pm$ 14.70 &            \phantom{0}67.19 $\pm$ 10.50 &        \phantom{0}61.80 $\pm$ \phantom{0}8.35 \\
\textbf{Medical Insurance (n=9)                  } &  \bftab\phantom{0}84.96 $\pm$ \phantom{0}0.00 &                      \phantom{0}56.47 $\pm$ 13.40 &            \phantom{0}48.05 $\pm$ 10.97 &                  \phantom{0}49.62 $\pm$ 10.79 \\
\textbf{Breast Cancer Wisconsin Original (n=9)             } &  \bftab\phantom{0}73.91 $\pm$ \phantom{0}0.00 &            \phantom{0}30.87 $\pm$ \phantom{0}2.25 &  \phantom{0}31.88 $\pm$ \phantom{0}0.00 &        \phantom{0}33.33 $\pm$ \phantom{0}7.45 \\
\textbf{Wisconsin Diagnostic Breast Cancer (n=10)} &  \bftab\phantom{0}82.14 $\pm$ \phantom{0}0.00 &                      \phantom{0}48.57 $\pm$ 22.64 &  \phantom{0}28.57 $\pm$ \phantom{0}0.00 &        \phantom{0}28.57 $\pm$ \phantom{0}0.00 \\
\textbf{Heart Disease (n=13)                     } &  \bftab\phantom{0}63.33 $\pm$ \phantom{0}0.00 &            \phantom{0}60.00 $\pm$ \phantom{0}0.00 &  \phantom{0}60.00 $\pm$ \phantom{0}0.00 &        \phantom{0}60.00 $\pm$ \phantom{0}0.00 \\
\textbf{Cervical Cancer (n=15)                   } &        \phantom{0}83.53 $\pm$ \phantom{0}0.00 &                      \phantom{0}77.06 $\pm$ 20.70 &  \phantom{0}84.59 $\pm$ \phantom{0}0.35 &  \bftab\phantom{0}84.71 $\pm$ \phantom{0}0.00 \\
\textbf{Loan House (n=16)                        } &            \bftab100.00 $\pm$ \phantom{0}0.00 &            \phantom{0}13.11 $\pm$ \phantom{0}0.00 &  \phantom{0}13.11 $\pm$ \phantom{0}0.00 &        \phantom{0}13.11 $\pm$ \phantom{0}0.00 \\
\textbf{Credit Card Default (n=23)                       } &  \bftab\phantom{0}75.03 $\pm$ \phantom{0}0.00 &                      \phantom{0}65.25 $\pm$ 16.47 &  \phantom{0}74.73 $\pm$ \phantom{0}0.00 &                  \phantom{0}65.57 $\pm$ 16.60 \\
\midrule
\textbf{Mean Fidelity                                  } &                  \phantom{0}82.30  &                      \phantom{0}49.89  &            \phantom{0}51.02 &                  \phantom{0}49.59 \\
\bottomrule
\end{tabular}
}
\caption{\textbf{Real-World Evaluation Results for Univariate Soft Decision Trees.}}
\label{tab:eval-results-sdt1}
\end{table}

Finally, for standard SDTs, the \mbox{$\mathcal{I}$-Net} achieved a significantly higher fidelity on $3/8$ datasets. The best sampling strategy for SDTs was a standard uniform distribution, which achieved a significantly higher accuracy than the \mbox{$\mathcal{I}$-Net} on $2/8$ datasets. Comparing the mean fidelities, the performance of the \mbox{$\mathcal{I}$-Net} with a fidelity of $92.1\%$ was considerably higher than sampling from a standard uniform distribution ($82.2\%$). This was mainly caused by the superior performance of the \mbox{$\mathcal{I}$-Net} on the \emph{Cervical Cancer} and \emph{Credit Card Default} dataset, where the fidelity was more than $40$ percentage points higher than sampling from standard uniform distribution. \\

\begin{table}[tb]
\centering
\resizebox{\columnwidth}{!}{
\begin{tabular}{l|cccc}
\toprule
\textbf{Dataset} &                                          \mbox{\textbf{\boldsymbol{$\mathcal{I}$}-Net}} & \textbf{Multi-Distribution} &                         \textbf{Standard Uniform} &                                \textbf{Standard Normal} \\
\midrule
\textbf{Titanic (n=9)                            } &  \bftab\phantom{0}95.51 $\pm$ \phantom{0}0.00 &            \phantom{0}88.31 $\pm$ \phantom{0}3.80 &        \phantom{0}92.47 $\pm$ \phantom{0}1.24 &  \phantom{0}92.81 $\pm$ \phantom{0}0.75 \\
\textbf{Medical Insurance (n=9)                  } &        \phantom{0}77.44 $\pm$ \phantom{0}0.00 &                      \phantom{0}79.25 $\pm$ 20.79 &  \bftab\phantom{0}91.20 $\pm$ \phantom{0}7.59 &  \phantom{0}78.50 $\pm$ \phantom{0}0.60 \\
\textbf{Breast Cancer Wisconsin Original (n=9)             } &                  100.00 $\pm$ \phantom{0}0.00 &            \phantom{0}96.67 $\pm$ \phantom{0}4.49 &                  100.00 $\pm$ \phantom{0}0.00 &  \phantom{0}31.88 $\pm$ \phantom{0}0.00 \\
\textbf{Wisconsin Diagnostic Breast Cancer (n=10)} &        \phantom{0}94.64 $\pm$ \phantom{0}0.00 &                      \phantom{0}84.82 $\pm$ 16.98 &  \bftab\phantom{0}97.50 $\pm$ \phantom{0}2.55 &  \phantom{0}28.57 $\pm$ \phantom{0}0.00 \\
\textbf{Heart Disease (n=13)                     } &                  100.00 $\pm$ \phantom{0}0.00 &                      \phantom{0}90.67 $\pm$ 12.45 &        \phantom{0}99.33 $\pm$ \phantom{0}1.33 &  \phantom{0}60.00 $\pm$ \phantom{0}0.00 \\
\textbf{Cervical Cancer (n=15)                   } &  \bftab\phantom{0}85.88 $\pm$ \phantom{0}0.00 &                      \phantom{0}58.35 $\pm$ 21.36 &                  \phantom{0}43.29 $\pm$ 11.54 &  \phantom{0}25.76 $\pm$ \phantom{0}2.38 \\
\textbf{Loan House (n=16)                        } &                  100.00 $\pm$ \phantom{0}0.00 &                      \phantom{0}50.82 $\pm$ 39.36 &                  100.00 $\pm$ \phantom{0}0.00 &            \phantom{0}17.21 $\pm$ 12.30 \\
\textbf{Credit Card Default (n=23)                       } &  \bftab\phantom{0}83.30 $\pm$ \phantom{0}0.00 &                      \phantom{0}59.86 $\pm$ 21.99 &        \phantom{0}38.77 $\pm$ \phantom{0}6.06 &  \phantom{0}75.40 $\pm$ \phantom{0}1.32 \\
\midrule
\textbf{Mean Fidelity                                  } &        \phantom{0}92.10  &                      \phantom{0}76.09  &                  \phantom{0}82.82  &            \phantom{0}51.27 \\
\bottomrule
\end{tabular}
}

\caption{\textbf{Real-World Evaluation Results for Standard Soft Decision Trees.}}
\label{tab:eval-results-sdt}
\end{table}

Furthermore, we observed that the average fidelity of standard SDTs is considerably higher than the fidelity of univariate SDTs and standard DTs. We can trace this back to the fact that SDTs have a significantly higher complexity, especially with an increasing number of variables, as shown in Section~\ref{sssec:inets_dts_soft}.
This can also explain why the performance difference between a sample-based distillation and the \mbox{$\mathcal{I}$-Net} is comparatively small for SDTs:
While using meaningful samples for querying the neural network is very crucial when the surrogate model has low complexity, it is less crucial if the surrogate model is more complex, making it less reliant on focusing on the most important information. Accordingly, it is less likely that relevant areas are neglected with an increasing complexity of the surrogate model. However, for interpretability, we are usually interested in surrogate models with a comparatively low complexity that are understandable for humans. In this scenario, $\mathcal{I}$-Nets substantially outperformed sample-based methods. \\

Summed up, the \mbox{$\mathcal{I}$-Net} outperformed a sample-based distillation on the majority of datasets when training data was not accessible for each type of surrogate model considered in this paper. In total, the \mbox{$\mathcal{I}$-Net} achieved a significantly higher fidelity in $15/24$ evaluated cases and only in $5/24$ cases a significantly lower fidelity.
Especially for surrogate models with low complexity, sample-based approaches are dependent on proper querying. Therefore, using the \mbox{$\mathcal{I}$-Net} in such scenarios can achieve a higher fidelity of the surrogate model. This can be crucial since wrong explanations can lead to wrong decisions, as we will evaluate more in-depth in Section~\ref{sssec:case_study_eval}.

\subsubsection{Ablation Study}\label{sssec:improvements_eval}
In Section~\ref{sssec:data_generation} we introduced an improved data generation method which should be more robust in a real-world scenario, since it considers multiple different distributions. In the following, we will compare our new data generation method with the data generation method introduced by \citet{marton2022explanations}, which generates data based on the function family of the surrogate model and considers only a single distribution. 
As shown in Table~\ref{tab:eval-results-data-gen}, we were able to outperform the standard data generation method in $7/8$ datasets for standard DTs, $8/8$ datasets for univariate SDT and $6/8$ datasets for the standard SDT. 
Summed up, the improved data generation method achieved a higher fidelity in  $21/24$ cases, while the old data generation only achieved a higher fidelity in $1/24$ cases.
Comparing the average performance over all datasets, we also observed a significant increase in the accuracy using the new data generation method of $\approx 36$ percentage points for standard DTs and $\approx 43$ percentage points for univariate SDTs. For standard SDTs, the difference in the mean fidelity was significantly smaller, with only $\approx 6$ percentage points. One explanation could be the higher complexity of the surrogate model for standard STDs, as already discussed in Section~\ref{sssec:real_world_eval}. 

\begin{table}[tb]
\small
\centering
\begin{tabular}{l|cc|cc|cc}
\toprule
\multicolumn{1}{c|}{\multirow{2}{*}{\textbf{Dataset}}} & \multicolumn{2}{c|}{\textbf{Standard DT}} & \multicolumn{2}{c|}{\textbf{Univariate SDT}} & \multicolumn{2}{c}{\textbf{Standard SDT}} \\
\multicolumn{1}{c|}{} & \multicolumn{1}{c}{\textbf{new}} & \multicolumn{1}{c|}{\textbf{old}} & \multicolumn{1}{c}{\textbf{new}} & \textbf{old} & \textbf{new} & \textbf{old} \\ \midrule

\textbf{Titanic (n=9)                            } &  \bftab\phantom{0}95.51 & \phantom{0}39.33 &  \bftab\phantom{0}95.51 & \phantom{0}60.67 &  \bftab\phantom{0}95.51 & \phantom{0}86.52 \\
\textbf{Medical Insurance (n=9)                  } &  \bftab\phantom{0}82.71 & \phantom{0}72.93 &  \bftab\phantom{0}84.96 & \phantom{0}77.44 &        \phantom{0}77.44 & \bftab\phantom{0}92.48 \\
\textbf{Breast Cancer Wisconsin Original (n=9)   } &  \bftab\phantom{0}97.10 & \phantom{0}31.88 &  \bftab\phantom{0}73.91 & \phantom{0}31.88 &  \bftab100.00           & \phantom{0}98.55 \\
\textbf{Wisconsin Diagnostic Breast Cancer (n=10)} &  \bftab\phantom{0}80.36 & \phantom{0}28.57 &  \bftab\phantom{0}82.14 & \phantom{0}28.57 &  \bftab\phantom{0}94.64 & \phantom{0}83.93  \\
\textbf{Heart Disease (n=13)                     } &  \bftab\phantom{0}73.33 & \phantom{0}60.00 &  \bftab\phantom{0}63.33 & \phantom{0}60.00 &  \bftab100.00           & \phantom{0}80.00 \\
\textbf{Cervical Cancer (n=15)                   } &  \phantom{0}84.71       & \phantom{0}84.71 &  \bftab\phantom{0}83.53 & \phantom{0}15.29 &  \bftab\phantom{0}85.88 & \phantom{0}84.71 \\
\textbf{Loan House (n=16)                        } &            \bftab100.00 & \phantom{0}13.11 &  \bftab100.00           & \phantom{0}13.11 &        100.00           & 100.00 \\
\textbf{Credit Card Default (n=23)                       } &  \bftab\phantom{0}75.80 & \phantom{0}74.73 &  \bftab\phantom{0}75.03 & \phantom{0}25.27 &  \bftab\phantom{0}83.30 & \phantom{0}64.97 \\
\midrule
\textbf{Mean Fidelity                            } &  \bftab\phantom{0}86.19 & \phantom{0}50.66 &  \bftab\phantom{0}82.30 & \phantom{0}39.03 &  \bftab\phantom{0}92.10 & \phantom{0}86.40 \\
\bottomrule
\end{tabular}
\caption{\textbf{Data Generation Performance Comparison.}}
\label{tab:eval-results-data-gen}
\end{table}


\subsubsection{Case Study: Explaining Neural Networks for Credit Card Default Prediction}\label{sssec:case_study_eval}
In this section, we will take a closer look at the explanations generated by sample-based approaches and the \mbox{$\mathcal{I}$-Net} by returning to \emph{Example~1} which we introduced in Section~\ref{sec:introduction}. 
The purpose of this experiment is to show in a real-world setting that without access to the training data, the surrogate model generated by sample-based approaches can lead to incorrect assumptions on the function learned by the neural network.
We want to note that without access to the training data, it is not possible to identify for a specific surrogate model whether it contains a misconception or not, since we are not able to calculate a representative fidelity.

The \emph{Credit Card Default} dataset is concerned with credit card default prediction based on $23$ features including demographic data and the credit history of clients in Taiwan~\citep{dataset_credit_card}. The demographic features include the sex, the education level, the marital status and the age. Furthermore, the repayment status, the amount of the bill and the amount of previous payments, each for the past $6$ months and the credit limit are listed for each client. The dataset is available in the UCI Machine Learning repository~\citep{UCI_ml_repository}, and a more detailed summary of the features with a short explanation is given in Table~\ref{tab:dataset_description}. \\

\begin{figure}[tb]
\centering
\begin{subfigure}{.5\columnwidth}
  \centering
  \includegraphics[width=.99\columnwidth]{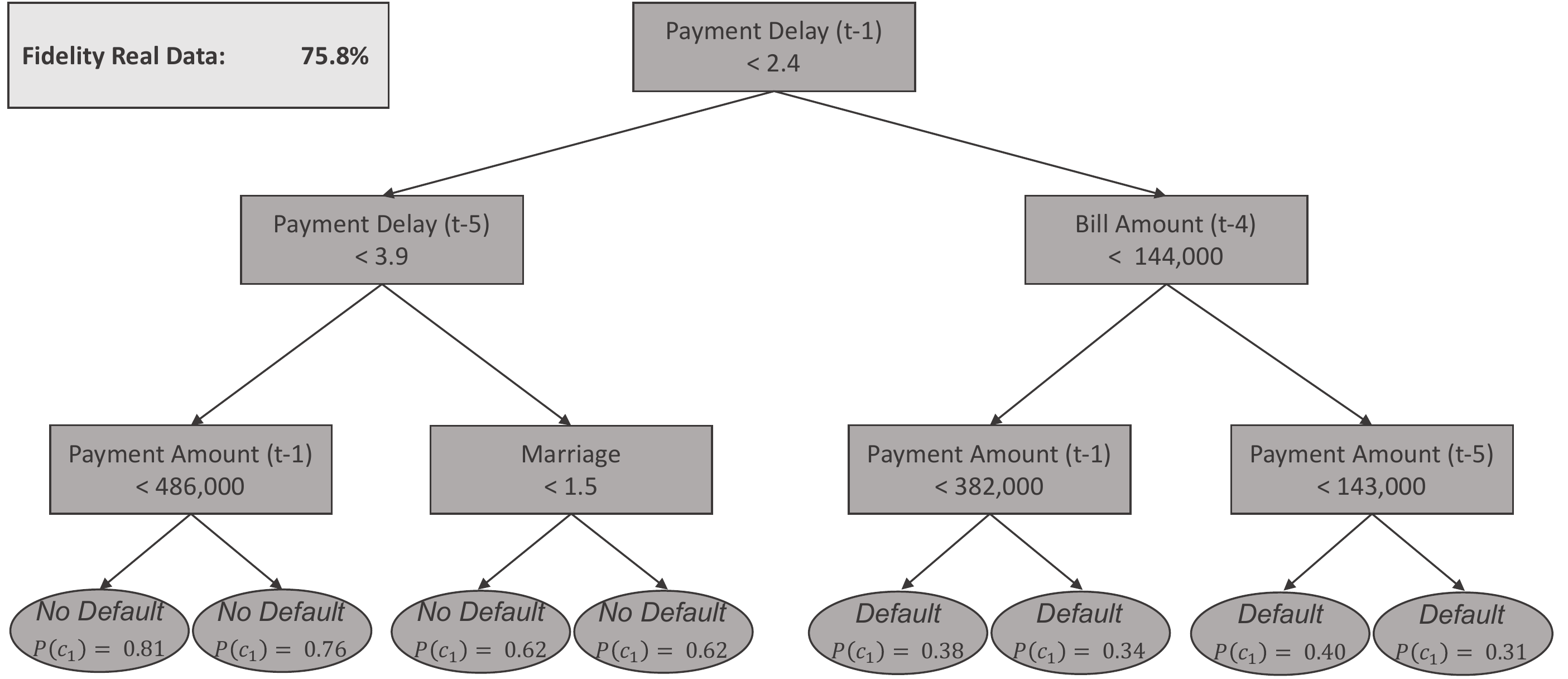}
  \caption{\mbox{$\mathcal{I}$-Net} Standard DT}
  \label{fig:vanilla_credit_card_inet}
\end{subfigure}%
\begin{subfigure}{.5\columnwidth}
  \centering
  \includegraphics[width=.99\columnwidth]{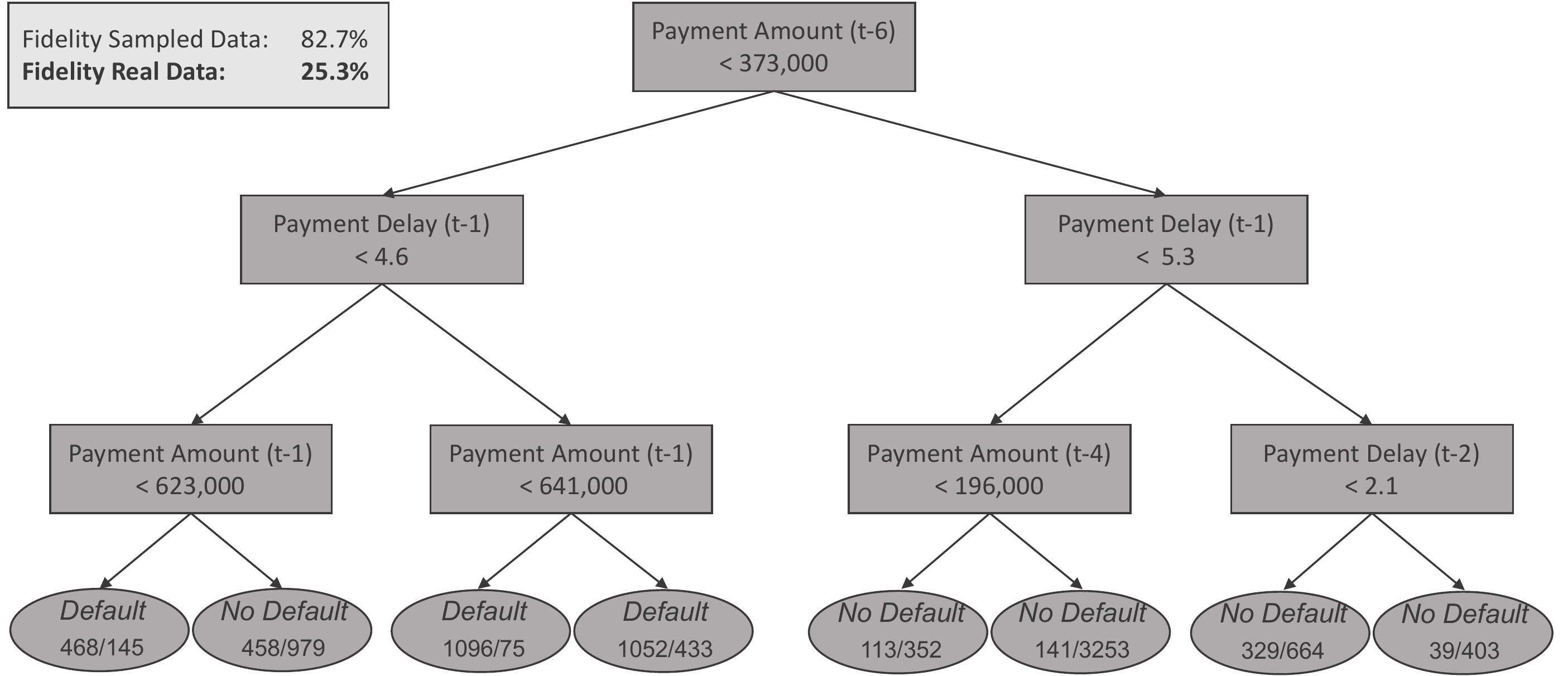}
  \caption{Sample-Based Standard DT}
  \label{fig:vanilla_credit_card_sample}
\end{subfigure}
\caption{\textbf{Explanation Comparison for Standard Decision Trees.} The surrogate model on the right is learned by a sample-based distillation with a multi-distribution sampling strategy. The DT on the left is predicted by the \mbox{$\mathcal{I}$-Net}. The \mbox{$\mathcal{I}$-Net} makes reasonable splits and achieves a significantly higher fidelity on the real data.}
\label{fig:vanilla_credit_card}
\end{figure}

Figure~\ref{fig:vanilla_credit_card} shows the standard DT surrogate models generated by the \mbox{$\mathcal{I}$-Net} and a sample-based distillation using the multi-distribution sampling strategy.
As shown in Figure~\ref{fig:vanilla_credit_card_inet}, the \mbox{$\mathcal{I}$-Net} archived a fidelity of $75.8\%$ and only considers a single split to decide whether there will be a payment or not. The split is based on whether the payment for the previous month was delayed for less than three months (left path) or not (right path). We can consider this as a very reasonable split, if we assume that if a client was in default previously, there is a higher possibility that there will be a default again. If we take the probabilities at the leafs into account, we can get some more information on the decision process. If the payment for 5 months ago was also delayed less than 4 months, the probability that there will be no default is even higher, as shown in the left branch of the tree. If there was a delay of more than 4 months, the probability that there will be a default is approximately $20\%$ higher. Furthermore, as we can see in the split at the bottom-right, if the past payment amount was higher than $143,000$, the chance of a default is approximately $10\%$ higher in this branch.

In contrast, when taking a closer look at the DT generated by a sample-based distillation (Figure~\ref{fig:vanilla_credit_card_sample}), we can observe that the entire right branch of the tree has \emph{No Default} as prediction. This prediction is made solely based on the first split, where the right branch is taken if the payment amount $6$ months ago was larger than $373,000$. This translates to the rule that we should always predict that there will be no default in the payment if there was a large payment amount in the past. However, it seems counter-intuitive to make this decision without taking for instance the credit history of the client and whether there were defaults previously into account. This is confirmed by the poor fidelity of the surrogate on the real data, which was only $25.3\%$ and worse than a random guess. However, the surrogate model had a very high fidelity of $82.7\%$ on the sampled data used for querying the model which leads to this misconception, since the model appears to have a high fidelity that does not hold on the real data. Without access to the training data, it is not possible to identify this misconception and taking the high fidelity on the sampled data into account, we might assume that the surrogate actually represents what the model has learned and therefore would make wrong assumptions on its behavior.

We can observe similar misconceptions for the function family of univariate and standard SDTs. However, we will not discuss them here in detail, but refer to the Appendix~\ref{A:case_study} for an in-depth evaluation of the explanations generated for those function families.

\section{Related Work}\label{sec:related_work}
Various methods to interpret black-box models have been proposed in the past decades. Overviews from different perspectives are given by \citet{doshi2017towards}, \citet{lipton2018mythos} and \citet{molnar2020interpretable}. In this paper, we focus on methods that translate neural networks into DTs to interpret the underlying decision function.

Model distillation is a common technique to transfer knowledge from a complex model into a surrogate model \citep{bucilua2006model,hinton2015distilling}. It can be used to obtain more compact model representations for efficiency reasons \citep{bucilua2006model,hinton2015distilling,furlanello2018born} or to interpret the model as a human-understandable function \citep{frosst2017distilling,tan2018learning}. 
With the focus on interpretability, model distillation is performed to either understand the function encoded by trained networks and how predictions are made \citep{craven1996extracting,boz2000converting,zhang2019interpreting} or to improve the performance of an interpretable algorithm to use it instead of the neural network at test time \citep{krishnan1999extracting,frosst2017distilling,liu2018improving}. Although those purposes differ, the methods can be interchangeably used for both. 


Various sample-based methods using DTs as surrogate models were presented in the past quarter-century \citep{craven1996extracting,krishnan1999extracting,boz2000converting,johansson2009evolving,frosst2017distilling,liu2018improving,zhang2019interpreting,nguyen2020towards}. 
These approaches have in common that they transform a trained neural network into a surrogate function with a tree-like structure. This is usually achieved by maximizing the fidelity to the neural network on a sample basis.
The main differences among existing approaches are the type of the resulting DTs, the method to train the surrogate model, and the purpose of the surrogate model.

The proposed trees make either univariate \citep{krishnan1999extracting,boz2000converting,liu2018improving}, $m$-of-$n$ \citep{craven1996extracting} or multivariate \citep{nguyen2020towards, frosst2017distilling} decisions at each split. 
Trees that consider multiple variables can achieve higher fidelity and accuracy than univariate DTs. However, especially for tabular data, the interpretation becomes harder.

For training the surrogate model, differences exist regarding the data used, the decision how a split is determined, and the optimization technique used.
Regarding the training of trees, most approaches rely on standard DT induction methods. \citet{krishnan1999extracting} use ID3 \citep{quinlan1986induction} and C4.5 \citep{quinlan2014c4}, \citet{craven1996extracting} use ID2-of-3 \citep{murphy1991id2} and \citet{nguyen2020towards} use CART \citep{breiman1984classification}. 
While these approaches greedily optimize the fidelity, \citet{frosst2017distilling} use gradient descent to distill the trees.

In the literature, the data to maximize the fidelity is either the training data used for the neural network \citep{frosst2017distilling,liu2018improving} or data from a distribution that was modeled based on the train data \citep{craven1996extracting,krishnan1999extracting,boz2000converting,johansson2009evolving}. 
The latter has been claimed to be an effective way to improve the results \citep{boz2000converting,johansson2009evolving}. 

In all cases, the predictions on data are the only source for understanding the black-box model behavior, and thus meaningful examples are crucial for the performance.
Without information about the distribution of the training data, e.g., in the form of data points, the performance of sample-based methods decreases significantly.
Recent model distillation approaches deal with this problem using metadata, such as layer activations, to create good samples based on network information \citep{lopes2017data,bhardwaj2019dream,nayak2019zero}. 
However, they often still need access to the training data in some part of the distillation process. \citet{lopes2017data} use a fraction of the original training data to compute activations summaries to later compress the network without accessing the data. Similarly, \citet{bhardwaj2019dream} require samples of the original training data to generate activation vectors, which are necessary for their distillation. However, they report requiring significantly fewer data points than \citet{lopes2017data}. In contrast, \citet{nayak2019zero} does not require access to training data, but only requires the model internals. The model internals are used to generate a class similarity matrix based on the parameters of the softmax output layer of the neural network. Based on the class similarity matrix, \citet{nayak2019zero} generate meaningful samples called \emph{Data Impressions} via Dirichlet sampling based on the output classes. However, the approach requires a softmax output for the neural network and is tailored towards multi-class classification problems, since it utilizes the knowledge contained in the class similarity matrix for sampling. Accordingly, an application on a binary classification task is not adequate, since a class similarity matrix for two classes can contain only little information, which makes sampling difficult. 
Summed up, the main issue is that the majority of state-of-the-art sample-free approaches still need access to at least a subset of the training data. Only \citet{nayak2019zero} is applicable if no training data is available, but the application is restricted, e.g., to multi-class tasks.

\section{Conclusion and Future Work}\label{sec:conclusion}
In this paper, we proposed a new instance of \emph{Interpretation Networks} (\mbox{$\mathcal{I}$-Nets}) for tree-based surrogate models and an improved data generation method, making \mbox{$\mathcal{I}$-Net} applicable in a real-world scenario. 
While traditional approaches generate explanations sample-based and therefore rely on proper querying, \mbox{$\mathcal{I}$-Nets} utilize the model internals, which implicitly contain all relevant information about the network function. Therefore, \mbox{$\mathcal{I}$-Nets} can generate reasonable explanations in scenarios where the training data is not accessible.


Using our new data generation method, we allow the \mbox{$\mathcal{I}$-Net} to learn how to generalize to neural networks trained on different data distributions. Thereby, the \mbox{$\mathcal{I}$-Net} identifies which aspects learned by the neural network are most important based on the distribution of the training data and therefore should be contained in the explanation. The \mbox{$\mathcal{I}$-Net} can use this knowledge to generate meaningful explanations for new, unseen networks, even without access to the training data.

In our experiments, we showed that sample-based approaches strongly rely on proper querying and are often not able to generate reasonable explanations once they have no access to the training data. In this scenario, the explanations of sample-based approaches frequently comprise misconceptions, since they focus on the global behavior and do not focus on the regions that are important for a reasonable explanation, as we demonstrated within our case study.
Furthermore, we empirically showed that \mbox{$\mathcal{I}$-Nets} consistently outperform sample-based methods on real-world datasets when the training data is not available. 
Thus, using \mbox{$\mathcal{I}$-Nets}, high-fidelity explanations can be generated when confidential training data can't be exposed.

Currently, the \mbox{$\mathcal{I}$-Net} comprises a feed-forward neural network and the model internals used as an input are flattened to a one-dimensional array. In further work, we aim for a more sophisticated \mbox{$\mathcal{I}$-Net} architecture and a better-suited representation for the model input to improve the performance even further.
Furthermore, the \mbox{$\mathcal{I}$-Net} is tailored towards generating fully grown DTs based on its structure. In further work this could be addressed by adjusting the output layer which allows using greater depths for the explanation without a significant increase in complexity.

\appendix

 \section{Hyperparameter Summary}\label{A:hyperparams}%

The hyperparameters for the $\mathcal{I}$-Net (Table~\ref{tab:i-net-params}) were tuned using a greedy neural architecture search according to \citet{jin2019auto}, followed by a manual fine-tuning of the selected values. To measure the performance during the optimization, we used the validation loss on a distinct validation set $\Theta_\lambda$ comprising $1000$ network parameters.

\begin{table}[ht]
\small
\centering
\begin{threeparttable}
\begin{tabular}{@{}cc|c@{}}
\toprule
\multicolumn{2}{c|}{\textbf{Parameter}} & \textbf{Value} \\ \midrule
\multirow{3}{*}{DT} & Hidden Layer Neurons  & \([1792, 512, 512]\)  \\ 
 & Hidden Layer Activation  & Sigmoid  \\
 & Dropout & \([0, 0, 0.5]\) \\ \midrule
 
\multirow{3}{*}{\begin{tabular}[c]{@{}c@{}}Univariate\\ SDT\end{tabular}} & Hidden Layer Neurons  & \([4096, 2048]\)  \\ 
 & Hidden Layer Activation  & Swish\tnote{a}  \\
 & Dropout & \([0, 0.5]\) \\ \midrule
 
\multirow{3}{*}{\begin{tabular}[c]{@{}c@{}}Standard\\ SDT\end{tabular}} & Hidden Layer Neurons  & \([1792, 512, 512]\)  \\ 
 & Hidden Layer Activation  & Swish\tnote{a}  \\
 & Dropout & \([0.3, 0.3, 0.3]\) \\ \midrule

\multicolumn{2}{c|}{Batch Size} & 256\\ %
\multicolumn{2}{c|}{Optimizer} & Adam \\
\multicolumn{2}{c|}{Learning Rate} & 0.001 \\
\multicolumn{2}{c|}{Loss Function} & $\mathcal{L}_{\mathcal{I}\text{-Net}}$  \\
\multicolumn{2}{c|}{Training Epochs} & 500  \\ 
\multicolumn{2}{c|}{Early Stopping} & Yes  \\ \midrule
\multicolumn{2}{c|}{Number of Training Samples} & 9,000 \\ 
\bottomrule
\end{tabular}
\begin{tablenotes}
\item[a] \footnotesize The Swish activation function proposed by \citet{swish_activation} is defined by $swish(x) = x \times sigmoid(x)$ and is claimed to consistently match or outperform a ReLU activation.
\end{tablenotes}
\end{threeparttable}
\caption{\textbf{  $\boldsymbol{\mathcal{I}}$-Net Training Parameters.}}
\label{tab:i-net-params}
\end{table}

\begin{table}[H]
\small
\centering
\begin{threeparttable}
\begin{tabular}{@{}c|cccc@{}}
\toprule
\textbf{Parameter} & \textbf{Value}  \\ \midrule

Hidden Layer Neurons  & \([128]\)  \\ 
Hidden Layer Activation  & ReLU  \\
Dropout & No \\
Batch Size & 64\\ %
Optimizer & Adam \\
Learning Rate & 0.001 \\
Loss Function & binary\_crossentropy  \\
Training Epochs & 1,000  \\ 
Early Stopping & Yes  \\ \midrule
Number of Training Samples & 5,000 \\ 
\bottomrule
\end{tabular}
\end{threeparttable}
\caption{\textbf{\boldsymbol{$\lambda$}-Net Training Parameters.}}
\label{tab:lambda-net-params}
\end{table}

\begin{table}[H]
\small
\centering
\begin{threeparttable}
\begin{tabular}{@{}c|cccc@{}}
\toprule
\textbf{Parameter} & \textbf{Value}  \\ \midrule
max\_depth & 3 \\ \midrule

criterion & gini \\
min\_samples\_split & 2 \\
min\_samples\_leaf & 1 \\

\bottomrule
\end{tabular}
\end{threeparttable}
\caption{\textbf{Standard DT Training Parameters.}}
\label{tab:dt-params}
\end{table}

\begin{table}[H]
\small
\centering
\begin{threeparttable}
\begin{tabular}{@{}c|cccc@{}}
\toprule
\textbf{Parameter} & \textbf{Value}  \\ \midrule
depth & 3 \\ \midrule
learning\_rate & 0.01 \\
criterion & binary\_crossentropy \\

lambda & 0.001 \\
beta & 1 \\
weight\_decaly & 0.0005 \\

maximum\_path\_probability & True \\

\bottomrule
\end{tabular}
\end{threeparttable}
\caption{\textbf{SDT Training Parameters.}}
\label{tab:sdt-params}
\end{table}

\section{Real-World Dataset Specification}\label{A:real-world-data}%

A specification of the datasets along with the source of the dataset and the performance of the neural network that was learned can be found in Table~\ref{tab:datasets}. A specification of the  hyperparameters for learning the neural networks can be found in Table~\ref{tab:lambda-net-params}.

For the \emph{Medical Insurance} dataset, the original objective is to predict the individual medical cost charged by the insurance. We transformed this to a classification task with the objective of predicting whether the medical cost is greater than $10000\$$ or not. 
The \emph{Heart Disease} dataset originally contains $75$ attributes. However, only $13$ are commonly used in published experiments. Therefore, we similarly only used these $13$ features. Furthermore, We distinguish only between \emph{Presence} (values $1,2,3,4$) and \emph{Absence} (value $0$), as it is common practice.
For the \emph{Cervical Cancer} dataset, we selected relevant features similar to the experiments conducted by \citet{molnar2020interpretable} and \citet{dataset_cervical_cancer}.

For the remainder of the datasets, no feature selection or specific transformation was performed. We used standard preprocessing for all datasets which includes the following steps:
\begin{enumerate}
    \item Remove all features comprising identifier features (e.g., IDs, Names).
    \item Impute missing values: For numeric values we used the mean for imputation, for ordinal, categorical and nominal features, we used the mode.
    \item Transform ordinal features to numeric values.
    \item One-hot-encode categorical and nominal features.
    \item Scale features in $[0, 1]$ using min-max normalization.
    \item Split data into distinct train ($85\%$), valid ($5\%$) and test set ($10\%$).
    \item Rebalance train data if number of samples of minority class is less than $25\%$.
\end{enumerate}

\begin{table}[ht]
\centering
\resizebox{\columnwidth}{!}{
\begin{tabular}{l|C{3.1cm}C{3cm}lL{6.7cm}|C{2cm}}
\toprule
\textbf{Dataset}  & \textbf{Number of Features Preprocessed (Raw)} & \textbf{Number of Samples (True/False)} & \textbf{Citation} & \textbf{Source} & \textbf{Network Performance} \\
\midrule

\textbf{Titanic} & 9 (12) & 891 (342/549) & - & \url{https://www.kaggle.com/c/titanic/data} & 83.15 \\

\textbf{Medical Insurance} & 9 (7) & 1338 (626/712) & \citet{dataset_medical_insurance} & \url{https://www.kaggle.com/datasets/mirichoi0218/insurance} & 95.49  \\

\begin{tabular}[l]{@{}l@{}}\textbf{Brest Cancer} \\ \textbf{Wisconsin}\end{tabular} & 9 (10) &  699 (241/458) & \begin{tabular}[l]{@{}l@{}}\citet{UCI_ml_repository} \\ \citet{dataset_breast_cancer_original} \end{tabular} & \url{https://archive.ics.uci.edu/ml/datasets/Breast+Cancer+Wisconsin+\%28Original\%29} & 97.10 \\

\begin{tabular}[l]{@{}l@{}}\textbf{Wisconsin Diagnostic} \\ \textbf{Breast Cancer}\end{tabular} & 10 (10) & 569 (212/357) & \citet{UCI_ml_repository} & \url{https://archive.ics.uci.edu/ml/datasets/Breast+Cancer+Wisconsin+\%28Diagnostic\%29} & 98.21 \\

\textbf{Heart Disease} & 13 (65) & 303 (164/139) & \begin{tabular}[l]{@{}l@{}}\citet{UCI_ml_repository} \\ \citet{dataset_heart_disease} \end{tabular} & \url{https://archive.ics.uci.edu/ml/datasets/heart+disease} & 93.33 \\

\textbf{Cervical Cancer} & 15 (36) & 858 (55/803) & \begin{tabular}[l]{@{}l@{}}\citet{UCI_ml_repository} \\ \citet{dataset_cervical_cancer} \end{tabular}  & \url{https://archive.ics.uci.edu/ml/datasets/Cervical+cancer+\%28Risk+Factors\%29} & 84.71 \\

\textbf{Loan House} & 16 (12) & 614 (422/192) & - & \url{https://datahack.analyticsvidhya.com/contest/practice-problem-loan-prediction-iii/} & 77.05 \\

\textbf{Credit Card Default} & 23 (23) & 30000 (23364/6636) & \begin{tabular}[l]{@{}l@{}}\citet{UCI_ml_repository} \\ \citet{dataset_credit_card} \end{tabular} & \url{https://archive.ics.uci.edu/ml/datasets/default+of+credit+card+clients} & 78.30 \\

\bottomrule
\end{tabular}
}
\caption{\textbf{Dataset Specifications.} This table shows the dataset specifications, including the number of features and the number of samples for each class, along with the accuracy of the neural network that was learned. Furthermore, we list the source of the corresponding datasets. All datasets were accessed last on  15.05.2022.}
\label{tab:datasets}
\end{table}

\section{Case Study: Credit Card Default Prediction}\label{A:case_study}

\subsection{Credit Card Default Dataset Description}\label{Asub:dataset_description}

In Table~\ref{tab:dataset_description} we shortly describe the features of the \emph{Credit Card Default} dataset along with the feature index and name as given by \citet{dataset_credit_card}.

\begin{table}[ht]
\centering
\resizebox{0.9\columnwidth}{!}{
\begin{threeparttable}
\begin{tabular}{L{0.1\columnwidth}L{0.2\columnwidth}|L{0.7\columnwidth}}
\toprule
\textbf{Feature Index} & \textbf{Feature Name} & \textbf{Explanation}  \\ \midrule
0  & LIMIT\_BAL &Amount of the given credit (NT dollar): it includes both the individual consumer credit and his/her family (supplementary) credit. \\ 
1  & SEX & Gender (1 = male; 2 = female) \\ 
2  & EDUCATION & Education (1 = graduate school; 2 = university; 3 = high school; 4 = others) \\ 
3  & MARRIAGE & Marital status (1 = married; 2 = single; 3 = others) \\ 
4  & AGE & Age (year) \\ 
5  & PAY\_0 & Repayment status in September, 2005. The measurement scale for the repayment status is: 1 = pay duly; 1 = payment delay for one month; 2 = payment delay for two months; . . .; 8 = payment delay for eight months; 9 = payment delay for nine months and above. \\ 
6  & PAY\_2 & Repayment status in August, 2005 (scale same as above) \\ 
7  & PAY\_3 & Repayment status in July, 2005 (scale same as above) \\ 
8  & PAY\_4 & Repayment status in June, 2005 (scale same as above) \\ 
9  & PAY\_5 & Repayment status in May, 2005 (scale same as above) \\ 
10  & PAY\_6 & Repayment status in April, 2005 (scale same as above) \\ 
11  & BILL\_AMT1 & Amount of bill statement in September, 2005 (NT dollar) \\ 
12  & BILL\_AMT2 & Amount of bill statement in August, 2005 (NT dollar) \\ 
13  & BILL\_AMT3 & Amount of bill statement in July, 2005 (NT dollar) \\ 
14  & BILL\_AMT4 & Amount of bill statement in June, 2005 (NT dollar) \\ 
15  & BILL\_AMT5 & Amount of bill statement in May, 2005 (NT dollar) \\ 
16  & BILL\_AMT6 & Amount of bill statement in April, 2005 (NT dollar) \\ 
17 & PAY\_AMT1 & Amount of previous payment in September, 2005 (NT dollar)  \\ 
18  & PAY\_AMT2 & Amount of previous payment in August, 2005 (NT dollar) \\ 
19  & PAY\_AMT3 & Amount of previous payment in July, 2005 (NT dollar) \\ 
20  & PAY\_AMT4 & Amount of previous payment in June, 2005 (NT dollar) \\ 
21  & PAY\_AMT5 & Amount of previous payment in May, 2005 (NT dollar) \\ 
22  & PAY\_AMT5 & Amount of previous payment in April, 2005 (NT dollar) \\ 
target  & target & Default payment (0=yes, 1=no) \\
\bottomrule
\end{tabular}
\end{threeparttable}
}
\caption{\textbf{Credit Card Dataset Feature Description.} The description of the feature is based on \citet{dataset_credit_card}.}
\label{tab:dataset_description}
\end{table}

\subsection{Explaining Neural Networks using Soft Decision Trees}\label{Asub:explanations_continued}

\begin{figure}[ht]
\centering
\begin{subfigure}{.5\columnwidth}
  \centering
  \includegraphics[width=.95\columnwidth]{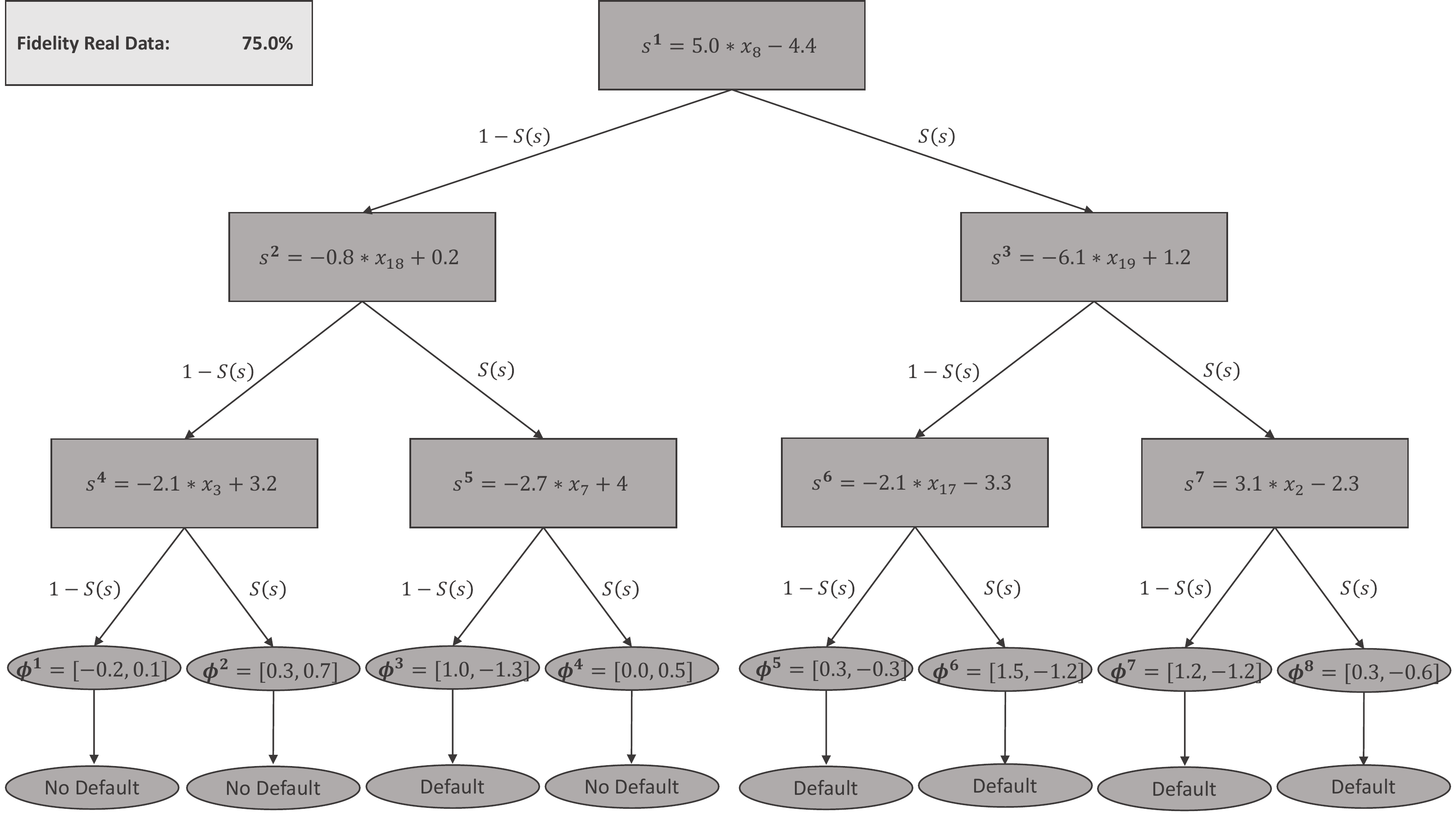}
  \caption{\mbox{$\mathcal{I}$-Net} Univariate SDT}
  \label{fig:sdt1_credit_card_inet}
\end{subfigure}%
\begin{subfigure}{.5\columnwidth}
  \centering
  \includegraphics[width=.95\columnwidth]{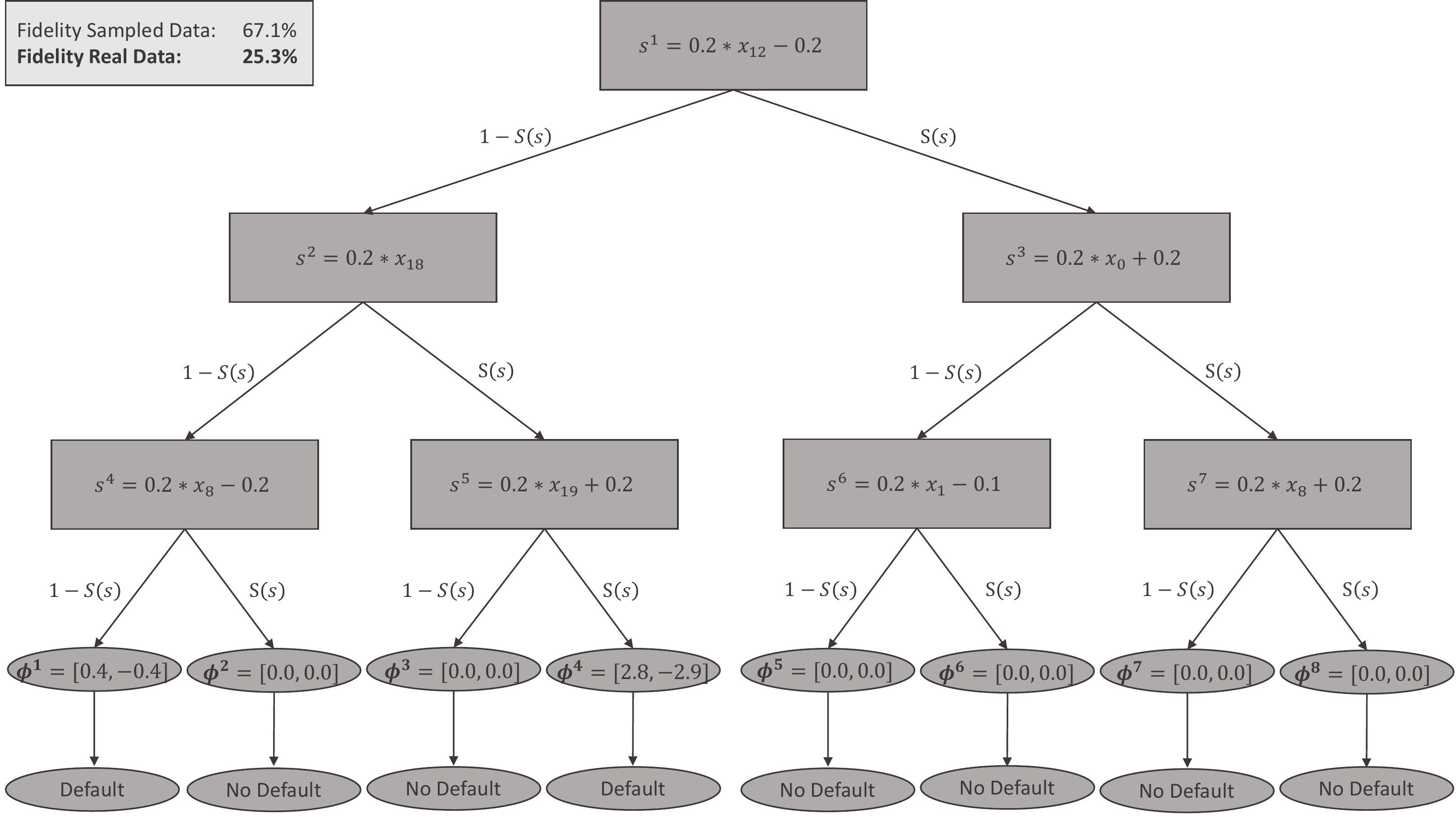}
  \caption{Sample-Based Univariate SDT}
  \label{fig:sdt1_credit_card_sample}
\end{subfigure}
\caption{\textbf{Explanation Comparison for Univariate Soft Decision Trees.} The surrogate model on the right is learned by a sample-based distillation with a multi-distribution sampling strategy. The tree on the left is predicted by the \mbox{$\mathcal{I}$-Net}. The \mbox{$\mathcal{I}$-Net} makes reasonable splits and achieves a significantly higher fidelity on the real data.}
\label{fig:sdt1_credit_card}
\end{figure}

\paragraph{Univariate Soft Decision Trees}
In Figure~\ref{fig:sdt1_credit_card}, we can see explanations for a neural network trained on the \emph{Credit Card Default} with univariate SDTs as surrogate model. There are no hard splits in a SDT, but we can still interpret the surrogate model by inspecting the internal nodes $s^i$. 
Thereby, a positive filter value translates into a higher probability of taking the right branch when increasing the corresponding input value.

The univariate SDT predicted by the \mbox{$\mathcal{I}$-Net} achieved a fidelity of $75\%$ which is similar to the fidelity of the standard DT. It considers the feature $x_8$ in the first internal node $s^1$ which corresponds to the repayment status $5$ months earlier. Since we can assume a high correlation between the repayment status for the different months and the target variable, we can consider this as a similar decision in the standard DT predicted by the \mbox{$\mathcal{I}$-Net}. A high value for this feature leads to an increased probability of taking the right path where a default of the payment is the most probable outcome in each leaf, which we can again consider as a reasonable explanation for this decision. 


For the univariate SDT generated by a sample-based distillation, there are many leafs where a parameter of $0$ was learned for both classes, which translates in a prediction of \emph{No Default}, but only with $50\%$ certainty. Additionally, the repayment status is not considered in neither of the first three internal nodes. If the right branch is taken with a higher probability, the univariate SDT always predicts that there will be no default. For this decision, only the bill amount $3$ months ago is considered. Similar to the standard DT generated by the sample-based distillation, this decision is made without considering the repayment status at all. Again, the fidelity on the sampled data with $67.1\%$ was significantly higher than the performance on the real data with only $25.3\%$. Taking this model to get insights to the functioning of the model would again lead to strong misconceptions.

\begin{figure}[ht]
\centering
\begin{subfigure}{.5\columnwidth}
  \centering
  \includegraphics[width=.95\columnwidth]{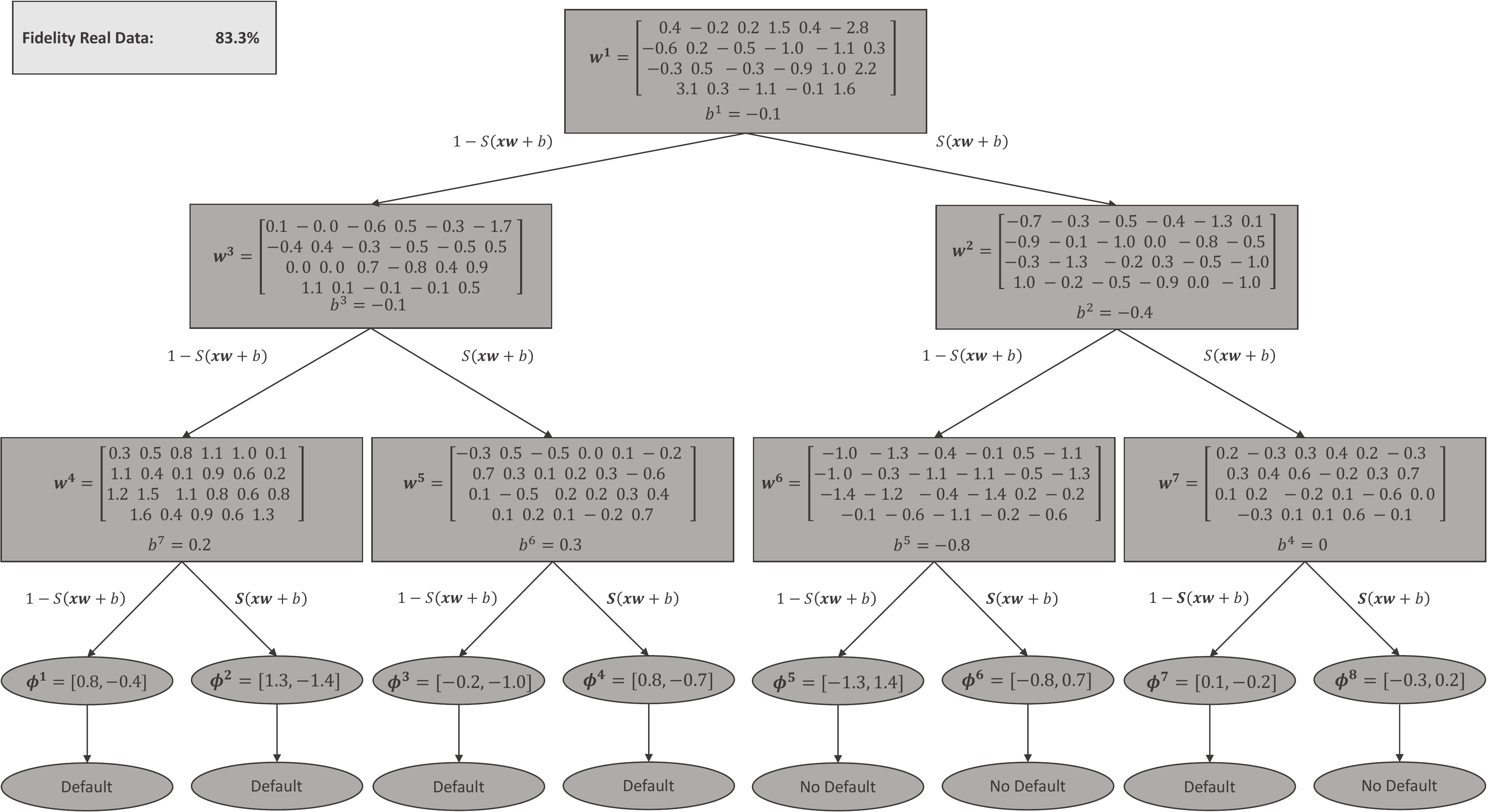}
  \caption{\mbox{$\mathcal{I}$-Net} Standard SDT}
  \label{fig:sdt_credit_card_inet}
\end{subfigure}%
\begin{subfigure}{.5\columnwidth}
  \centering
  \includegraphics[width=.95\columnwidth]{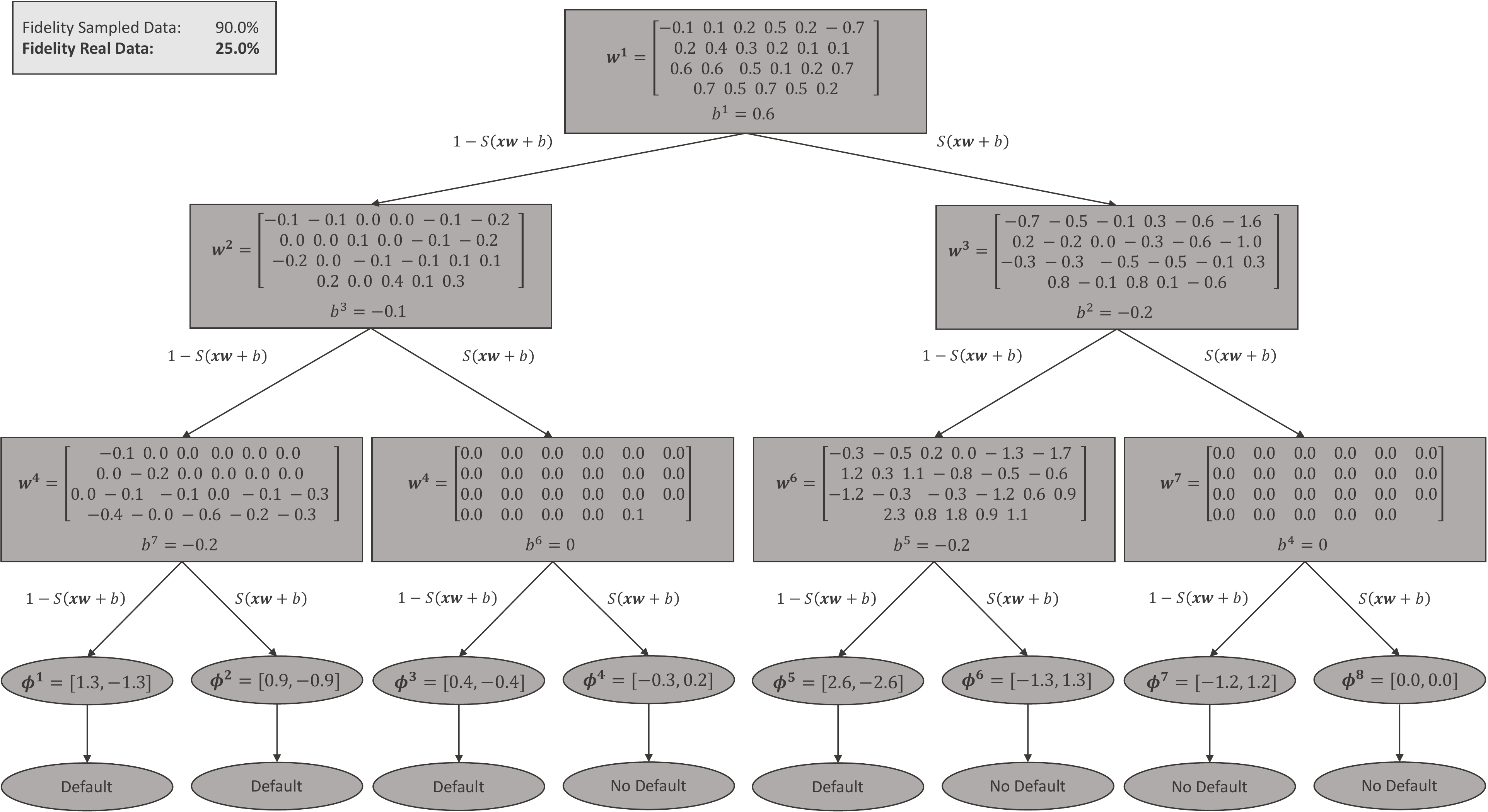}
  \caption{Sample-Based Standard SDT}
  \label{fig:sdt_credit_card_sample}
\end{subfigure}
\caption{\textbf{Explanation Comparison for Standard Soft Decision Trees.} The surrogate model on the right is learned by a sample-based distillation with a multi-distribution sampling strategy. The tree on the left is predicted by the \mbox{$\mathcal{I}$-Net}. The \mbox{$\mathcal{I}$-Net} makes reasonable splits and achieves a significantly higher fidelity on the real data.}
\label{fig:sdt_credit_card}
\end{figure}

\paragraph{Standard Soft Decision Trees}
Figure~\ref{fig:sdt_credit_card} shows standard SDTs as surrogate model. For the \mbox{$\mathcal{I}$-Net}, the fidelity of the standard SDTs was approximately $8$ percentage points higher than the fidelity of univariate DTs. However, this increase in the fidelity comes with a significant increase in the complexity, since each filter comprises $23$ values that are considered for calculating the probabilities of taking the left or right path. This makes standard SDTs much harder to comprehend for humans. However, we can still get some insights if we inspect the filters thoroughly.
In $\mathbf{w}_1$ of the SDT predicted by the \mbox{$\mathcal{I}$-Net} (Figure~\ref{fig:sdt_credit_card_inet}), the value at index $6$ and $18$ have a considerably higher absolute value than the remainder of the values, which makes them especially important for calculating the path probabilities. Accordingly, a high value for the repayment status of the last month (negative filter value) and a low value for the amount of the payment two months ago (positive filter value) strongly increases the probability of taking the left path, which results in a payment default as prediction. Therefore, the surrogate model assumes that most clients that recently had a default in their payment and recently had a low payment amount are likely to default again. However, this does not account for all clients, since the filter comprises values $\neq 0$ for all $23$ features. Accordingly, there exist data points where the probability of taking the right path is higher, even if there is a high value for the repayment status of the last month and a low value for the amount of the payment two months ago. This makes it increasingly hard to understand all aspects of the explanation, which is usually the goal when global surrogate models are selected as an explanation method.
The SDT generated by the sample-based distillation (Figure~\ref{fig:sdt_credit_card_sample}) is even more difficult to interpret. The filter $\mathbf{w}_1$ comprises many values with a similarly high absolute value and therefore many features have a similar importance, which makes it hard to formulate general explanations. It would be much easier to explain why there is a higher probability for taking a specific path for a certain sample. However, this is not the purpose of a global explanation, but in this instance it would be more reasonable to use a local explanation method. Furthermore, the surrogate model generated by a sample-based distillation again only achieved a high fidelity ($90\%$) on the sampled data, but a very low fidelity on the real data ($25\%$). Therefore, we can assume that the explanation is not able to give insights on the functioning of the neural network that hold in a real world scenario.

\section{Dataset Size for Sample-Based Distillation}

Selecting an appropriate number of data points to sample when using a sampling strategy to generate the query data is very crucial. However, with an increasing number of samples, the runtime also increases significantly. Figure~\ref{fig:inet_sdt} shows the mean performance of a sample-based distillation using standard DTs on the real-world datasets in Table~\ref{tab:datasets} for an increasing number of sample points. For each number of samples, we ran $10$ independent trials, similar to the experiments conducted in Section~\ref{sssec:real_world_eval}.

\begin{figure}[ht]
    \centering
    \includegraphics[width=0.70\columnwidth]{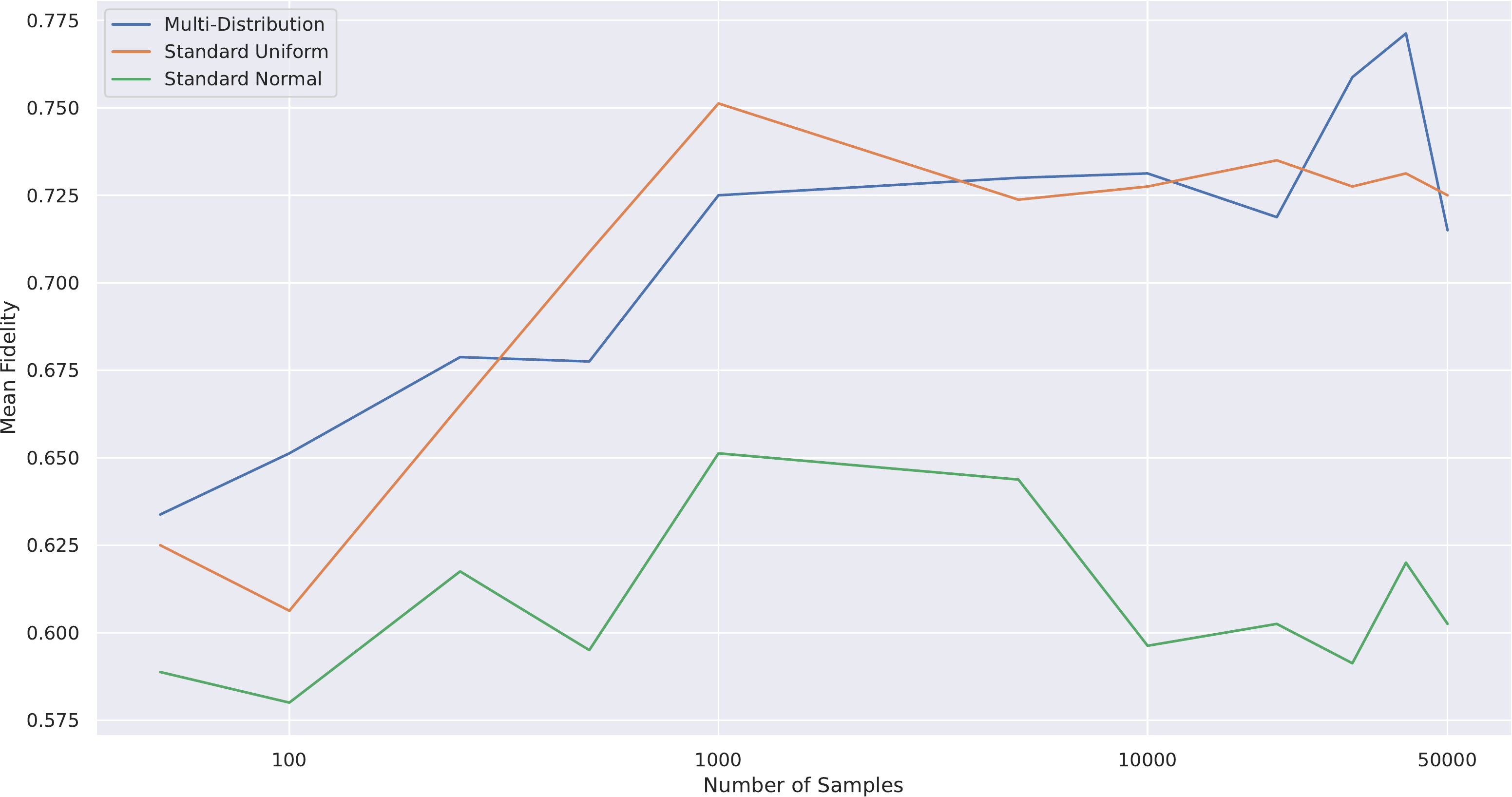}
    \caption{\textbf{Dataset Size for Sample-Based Distillation.} This figure shows the mean performance of sample-based approaches on the real-world datasets in Table~\ref{tab:datasets} when increasing the number of samples generated using the different sampling strategies. We can see that there is no considerable performance increase when increasing the number of samples above $10000$.}    
    \label{fig:sample_size}
\end{figure}


\bibliographystyle{unsrtnat}
\bibliography{references}  






\end{document}